\documentclass[10pt,twocolumn,letterpaper]{article}

\usepackage[pagenumbers]{cvpr} % To force page numbers, e.g. for an arXiv version

\definecolor{cvprblue}{rgb}{0.21,0.49,0.74}
\usepackage[pagebackref,breaklinks,colorlinks,allcolors=cvprblue]{hyperref}

\usepackage{colortbl,xcolor}
\definecolor{oursPink}{HTML}{FFB5A7}
\definecolor{oursGreen}{rgb}{0.82, 0.94, 0.86}
\definecolor{spotBlue}{RGB}{220,235,255}
\definecolor{failred}{HTML}{EB6B66}
\usepackage{multirow}
\usepackage{subcaption}
\usepackage{algorithm}
\usepackage{algpseudocode}
\usepackage{float}
\def\confName{CVPR}
\def\confYear{2027}

\title{When Slots Compete:\\
Slot Merging in Object-Centric Learning}

\author{
Christos Chatzisavvas$^{1,2,*}$ \quad
Panagiotis Rigas$^{3,4,*}$ \quad
Ioannis Chamodrakas$^{3}$\\
Nikolaos Mitianoudis$^{1,2}$ \quad
Vassilis Katsouros$^{2}$ \quad
George Ioannakis$^{2}$\\[0.5em]
{\small $^{1}$Department of Electrical and Computer Engineering,
Democritus University of Thrace, Greece}\\
{\small $^{2}$Institute for Language and Speech Processing,
Athena Research Center, Greece}\\
{\small $^{3}$Department of Informatics \& Telecommunications, National and Kapodistrian University of Athens, Greece}\\
{\small $^{4}$Archimedes, Athena Research Center, Greece}\\
{\small $^{*}$ Equal contribution}
}

\begin{document}
\maketitle
\begin{abstract}
Slot-based object-centric learning represents an image as a set of latent slots with a decoder that combines them into an image or features. The decoder specifies how slots are combined into an output, but the structure of the slot set is typically fixed: the number of slots is chosen upfront and slots are only refined. This can lead to multiple slots \emph{competing} for overlapping regions of the same entity rather than focusing on distinct regions. We introduce \emph{slot merging}: a drop-in, lightweight operation on the slot set that \emph{merges} overlapping slots during training. We quantify overlap with a \emph{Soft-IoU} score between slot-attention maps and combine selected pairs via a barycentric update that preserves gradient flow. Merging follows a fixed policy, with the decision threshold inferred from overlap statistics, requiring no additional learnable modules. Integrating into DINOSAUR, SPOT, and SlotDiffusion, slot merging improves object discovery and segmentation, achieving competitive performance compared with adaptive methods with only minor computational overhead. Its benefits extend to object recognition, set prediction, compositional generation, and slot-level scene editing.
\end{abstract}    
\section{Introduction}
\label{sec:intro}

Human perception is fundamentally linked to the ability to decompose visual environments into discrete, meaningful entities~\cite{peters2021capturing}. Object-Centric Learning (OCL) translates this cognitive principle into artificial systems by extracting structured, object-level representations from raw perceptual data. To reduce reliance on manual annotation, many approaches treat scene decomposition as an unsupervised reconstruction task. A prominent framework is Slot Attention~\cite{locatello2020object}, which serves as the architectural bottleneck enabling this decomposition. It operates on a set of $K$ latent variables, referred to as \emph{slots}, intended to encode individual objects or coherent scene components. Through an iterative, competitive attention mechanism, these slots aggregate information from spatial feature embeddings to explain distinct regions of a scene. The resulting representations can subsequently be decoded and used in a range of object-centric downstream tasks, including reasoning~\cite{chi2025slot}, classification~\cite{rubinstein2025we}, and manipulation~\cite{chapin2025object}.

\begin{figure}[t]
    \centering
    \includegraphics[width=0.93\linewidth, trim=20pt 10pt 20pt 10pt, clip]{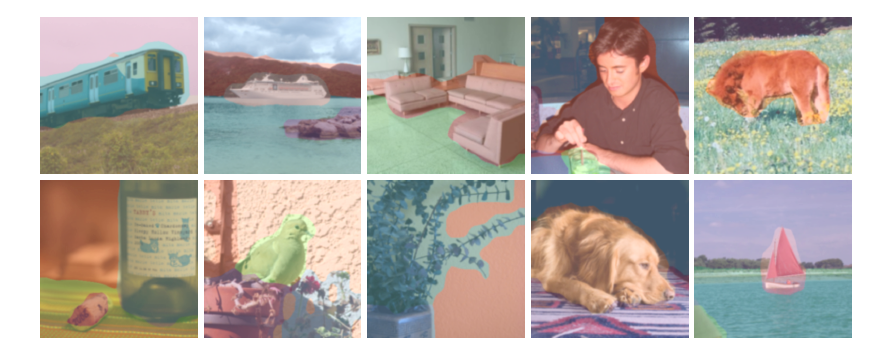}
    \caption{We introduce a merge operator over the slot set that adaptively refines factorization, producing coherent object-level representations.}
    \label{fig:segm_res}
\end{figure}

Recent advances have extended slot-based learning to complex, real-world scenes. DINOSAUR~\cite{seitzer2022bridging} learns object representations by reconstructing features, instead of pixels, from a pretrained visual encoder, while SPOT~\cite{Kakogeorgiou_2024_CVPR} further improves object discovery through an attention-based distillation, and patch-order permutation. These developments demonstrate that stronger visual features (e.g., DINO~\cite{caron2021emerging}) and training strategies (e.g., teacher-student) can substantially improve slot-based scene decomposition. However, the resulting slots do not always correspond to distinct objects~\cite{seitzer2022bridging}. Multiple slots may attend to overlapping regions of the same object, distributing its representation across several slots. This fragmentation can affect both segmentation quality and downstream tasks that rely on individual slots to represent complete objects.

We focus on this unresolved \emph{slot competition}, where multiple slots continue to explain the same object after Slot Attention has clustered the input features. Although competition encourages slots to specialize, the Slot Attention update rule does not explicitly combine their representations~\cite{locatello2020object}. Such slots may contain information about the same object, and can be brought together into a more coherent representation. This observation motivates a simple question: \textit{Can merging competing slots improve the object representations learned by existing slot-based models?}

To this end, we introduce \emph{slot merging}, a lightweight operation applied to the slots produced by Slot Attention before decoding. We identify candidate pairs using a probabilistic Intersection-over-Union (Soft-IoU) score between their attention maps. Selected pairs are combined through a mass-weighted convex combination (barycentric aggregation) of their representations, with the corresponding attention maps aggregated to reflect the merged slot. Pair selection follows a fixed rule, using a threshold estimated from the distribution of pairwise overlaps. The merge operator remains differentiable, allowing gradients to propagate to the contributing slots during training. The method requires no additional learnable modules, and can also be applied to pretrained models during inference. The resulting mechanism acts as a refinement of Slot Attention. We evaluate the proposed method within DINOSAUR~\cite{seitzer2022bridging}, SPOT~\cite{Kakogeorgiou_2024_CVPR}, and SlotDiffusion~\cite{wu2023slotdiffusion} against representative baselines and assess its benefits for downstream tasks. The overall pipeline is illustrated in Fig.~\ref{fig:merging_process}, and segmentation results are shown in Fig.~\ref{fig:segm_res}.

To summarize, our contributions are:
\begin{enumerate}
\item \textbf{Overlap-based slot merging.}
We introduce a slot-merging mechanism that identifies competing slots through the spatial overlap of their attention maps, with a data-driven threshold regulating the merging process.

\item \textbf{Differentiable merge operator.}
We combine competing slot representations through a barycentric aggregation, allowing gradients to propagate to the contributing slots during training.

\item \textbf{Integration into existing models.}
Slot merging requires no additional learnable modules and can be incorporated into existing slot-based models during training or applied to pretrained models at inference.

\item \textbf{Experimental evaluation.}
We evaluate slot merging within DINOSAUR and SPOT on four object-centric learning benchmarks, demonstrating improvements in object discovery and downstream tasks, including object recognition, set prediction, compositional generation, and slot-level scene editing.
\end{enumerate}
\section{Related Work}
\label{sec:relatedwork}

\noindent\textbf{Unsupervised Object-Centric Learning.}
Object-Centric Learning aims to represent complex scenes as compositions of individual objects~\cite{burgess2019monet}. Early approaches relied on sequential generative inference, discovering objects by iteratively attending to image regions and representing them through object-wise latent variables~\cite{eslami2016attend,kosiorek2018sequential}. Subsequent mixture-based autoencoding frameworks~\cite{burgess2019monet,greff2019multi,engelcke2020genesis} decompose scenes into latent object components with soft pixel-wise assignments, avoiding explicit box parametrization. Slot Attention~\cite{locatello2020object} introduced competitive attention over encoded features, iteratively refining a set of permutation-equivariant object-centric representations. Related approaches explore distributed representations and improved grouping mechanisms~\cite{lowe2023rotating,kim2023shepherding,lowecomplex}. Extensions to video exploit recurrent slot propagation, motion or depth cues, and temporal feature consistency~\cite{kipf2022conditional,elsayed2022savi++,aydemir2023self,zadaianchuk2023object}.

\noindent\textbf{Slot-Based Architectures.}
Recent methods improve slot-based models through their reconstruction targets, decoders, and training strategies. DINOSAUR~\cite{seitzer2022bridging} scales object discovery to real-world scenes by reconstructing semantic features from a pretrained visual encoder. Other approaches use discretized or vector-quantized visual representations~\cite{singh2021illiterate,zhao2025vector}. Autoregressive transformer-based decoders~\cite{singh2021illiterate,seitzer2022bridging,Kakogeorgiou_2024_CVPR} provide expressive reconstruction, while diffusion-based ones~\cite{jiang2023object,wu2023slotdiffusion} improve the fidelity of compositional image generation through iterative denoising conditioned on object slots. Beyond decoder design, SPOT~\cite{Kakogeorgiou_2024_CVPR} improves object discovery through attention-based distillation, which distills teachers' decoder attention masks into the student Slot Attention module. Recent work also explores complementary information from multiple encoder layers~\cite{Bock_2026_CVPR} or foreground--background guidance~\cite{tian2025pay}, to improve object binding. These methods demonstrate that improving individual components can benefit existing object-centric architectures. In contrast, our work introduces an operator that consolidates overlapping slots without requiring architectural changes or additional learnable modules.

\noindent\textbf{Operations on Representations.}
Operations that change the set of learned primitives have been explored in prototype learning, mixture models, and Gaussian representations~\cite{fritzke1994growing,ueda2000smem,kerbl20233d,rigas2025garlic}. Within object-centric learning, spectral clustering based on cosine similarity between slot vectors is used to combine predicted regions at inference~\cite{tian2025pay}, while DIAS~\cite{zhao2025slot} applies clustering followed by slot re-initialization. MetaSlot~\cite{liu2025metaslot} uses prototype-based deduplication, removing redundant slots before further refinement. Other approaches adapt the active slot set, through learned selection in AdaSlot~\cite{fan2024adaptive} and quality-guided selection in QASA~\cite{ouyang2026qasa}, while GENESIS-V2~\cite{engelcke2021genesis} infers a variable number of components through stick-breaking. These approaches demonstrate different ways of adapting object representations to the content of a scene. Unlike methods based on slot selection or prototype deduplication, we directly combine the information carried by spatially overlapping representations. We identify competing slots through attention-map overlap and combine their representations.
\section{Method}
\label{sec:method}

\noindent\textbf{Preliminaries.}
Slot Attention~\cite{locatello2020object} maps a set of $N$ input features
$X \in \mathbb{R}^{N \times d}$
to $K$ object-centric embeddings (slots)
$S \in \mathbb{R}^{K \times d_{slots}}$ 
through an iterative cross-attention mechanism. Conceptually, Slot Attention can be interpreted as a learned variant of soft \emph{k}-means clustering, where slots act as cluster centroids that iteratively group input features into object-level representations.
Slots are initialized from a learnable Gaussian distribution
$S^{(0)} \sim \mathcal{N}(\mu, \operatorname{diag}(\sigma^2))$, which allows each slot to serve as a flexible prototype that can bind to arbitrary objects. At each iteration $t = 0, 1, \dots, T-1$, slots compete to explain the input features via attention.
The attention weights between inputs and slots are computed as:
\(
A
=
\operatorname{softmax}\!\left(
{K_x Q_s^{\top}}/{\sqrt{d}}
\right)
\in \mathbb{R}^{N \times K}, 
\text{where }
Q_s = q(S^{(t)}),\;
K_x = k(X),\;
V_x = v(X).
\)

Here, $q(\cdot)$, $k(\cdot)$, and $v(\cdot)$ are learned linear maps. The softmax is applied across the slot dimension, ensuring that attention is normalized across slots for each patch: \( \sum_{k=1}^{K} A_{n,k} = 1, \ \forall n. \) Such normalization induces competition between slots, encouraging specialization across distinct regions or objects. For notational simplicity, batch dimension is omitted in the Slot Attention formulation above and introduced explicitly in Alg.~\ref{alg:slot_merging}. 

Using these attention weights, each slot gathers a weighted summary of the input features: 

\begin{equation}
\hat{S}
=
\left(
\frac{A_{i,j}}{\sum_{l=1}^{N} A_{l,j}}
\right)^{\!\top}
V.
\end{equation}

This slot-wise summary $\hat{S}$ is then combined with the previous slot state $S^{(t)}$ using a Gated Recurrent Unit (GRU), followed by layer normalization and a lightweight MLP to produce the refined slots:

\begin{equation}
S^{(t+1)}
=
\operatorname{MLP}\!\Big(
\operatorname{LayerNorm}\big(
\operatorname{GRU}(S^{(t)}, \hat{S})
\big)
\Big).
\end{equation}

After $T$ refinement iterations, the final slots $S^{(T)}$ serve as object-centric representations for downstream modules. Our proposed merging mechanism operates on these refined slots, further optimizing them during training.

\begin{figure*}
    \centering
    \includegraphics[width=0.93\linewidth]{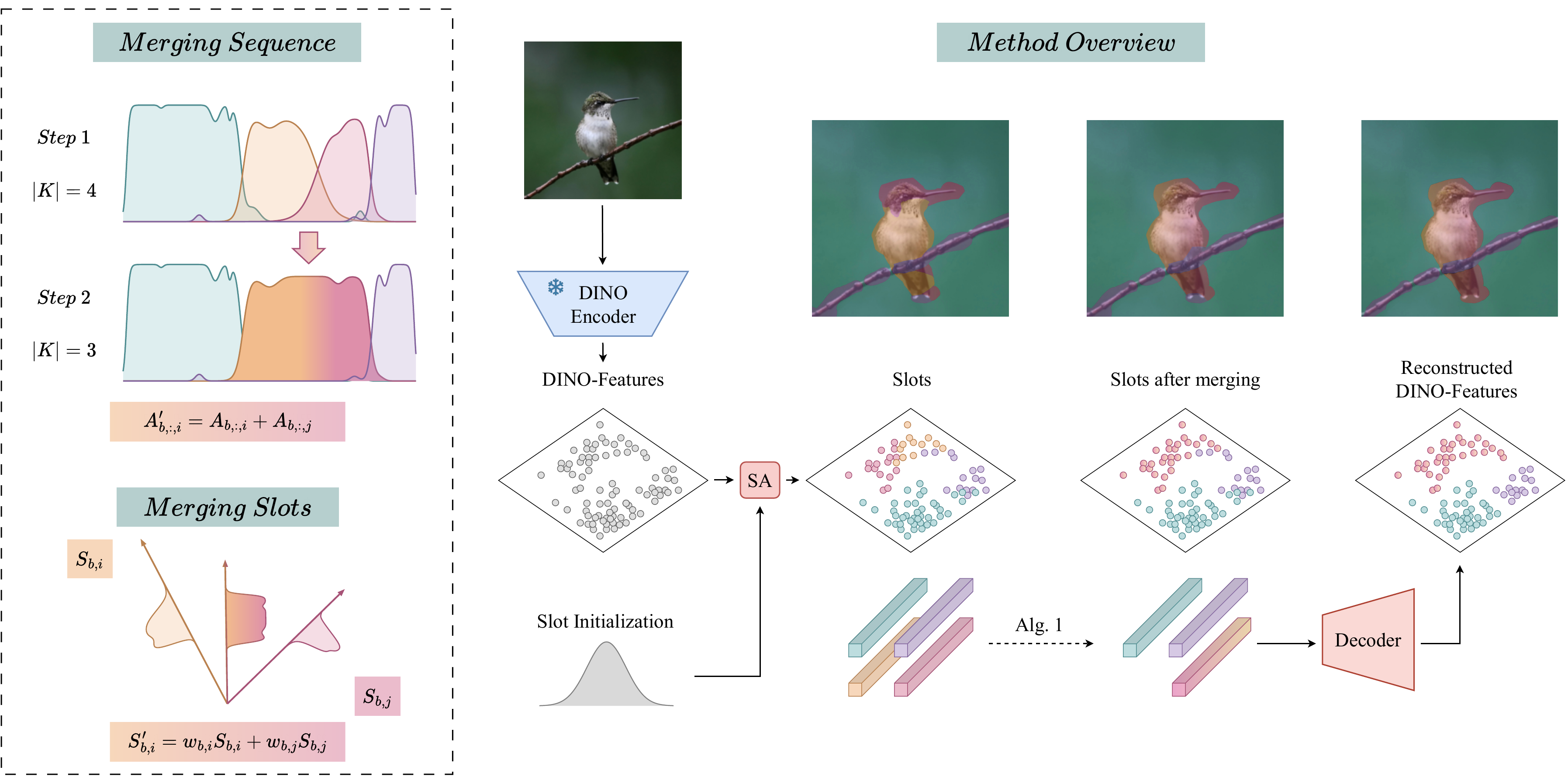}
    \caption{Overview of the proposed training pipeline and merging mechanism. After Slot Attention produces refined slots and attention maps, pairwise Attention-IoU scores are computed to identify the most overlapping slot pair. The selected slots are merged via barycentric aggregation and their attention maps are summed. The procedure is repeated iteratively until no overlap exceeds the threshold, yielding a refined slot set passed to the decoder.}
    \label{fig:merging_process}
\end{figure*}

\subsection{Quantifying Spatial Slot Competition}
\label{sec:iou-metric}
Effective slot merging requires identifying slots that correspond to the same underlying object. While slot vectors encode compact semantic representations, the spatial structure of their attention maps provides a more direct signal of object identity. When two slots attend to spatial locations with strong overlap, they compete to explain the same underlying structure, indicating an inefficient allocation of representational capacity.

Let $A\in[0,1]^{N\times K}$ denote the normalized attention matrix produced by Slot Attention, where $N$ is the number of spatial locations and $K$ the number of slots. The attention maps associated with slots $i$ and $j$ are $p=A_{:,i}$ and $q=A_{:,j}$ with $p,q\in[0,1]^N$. Interpreting these vectors as soft spatial masks, we quantify their overlap using a probabilistic variant of the Intersection-over-Union (IoU)~\cite{rahman2016optimizing}. The resulting similarity measure is defined as
\begin{equation}
\mathrm{IoU}(p,q)=
\frac{\sum_{n} p_n q_n}
{\sum_{n}\left(p_n+q_n-p_n q_n\right)} .
\label{eq:soft_iou}
\end{equation}
Let $I_{ij} = \mathrm{IoU}(A_{:,i}, A_{:,j})$ denote the pairwise overlap score between slots $i$ and $j$. The numerator represents the soft intersection, computed as the element-wise product of the two attention maps and summed over spatial locations. The denominator corresponds to the soft union, aggregating the spatial extent covered by each slot while subtracting the intersection term to avoid double counting. High IoU values therefore indicate substantial spatial agreement between slots, suggesting that they capture the same underlying entity. The resulting overlap scores are used to identify pairs of slots that compete for the same object and are therefore candidates for merging.

\subsection{Slot Merge Operator}
\label{sec:slot-mech}
Given two competing slots $i$ and $j$, identified through the overlap score defined in Sec.~\ref{sec:iou-metric}, we merge their representations using a mass-weighted operator. The operator combines both the slot vectors and their associated attention maps while preserving the total mass. Let $A\in[0,1]^{N\times K}$ denote the attention matrix and $S_i,S_j\in\mathbb{R}^D$ the corresponding slot representations. The attention mass of slot $i$ is defined as the sum of its attention weights over all spatial locations,
\begin{equation}
\alpha_i = \sum_{n=1}^{N} A_{n,i}.
\label{eq:mass}
\end{equation}
The relative contributions of the two slots are determined by normalized weights
\begin{equation}
w_i=\frac{\alpha_i}{\alpha_i+\alpha_j}, \qquad
w_j=\frac{\alpha_j}{\alpha_i+\alpha_j},
\label{eq:convweights}
\end{equation}
which satisfy $w_i+w_j=1$. These coefficients ensure that each slot contributes proportionally to its spatial support.
The merged slot representation is obtained through a barycentric update,
\begin{equation}
S_i \leftarrow w_i S_i + w_j S_j ,
\label{eq:barymerge}
\end{equation}
which replaces slot $i$ with the convex combination of the two slot vectors. 

The attention maps are updated accordingly by aggregation,
\begin{equation}
A_{:,i} \leftarrow A_{:,i} + A_{:,j}.
\label{eq:attnmapsupdate}
\end{equation}
After the update, slot $j$ is removed from the slot set. The resulting slot therefore preserves the total attention mass while aggregating the representations of the two competing slots. The merging operation is also illustrated in Fig.~\ref{fig:merging_process}

\subsection{Fixed Merge Policy}
\label{sec:ada-thres}

The merge operator defined in Sec.~\ref{sec:slot-mech} specifies how two slots are merged. A policy determines which slot pairs are merged and when the process terminates.
Given the pairwise overlap scores defined in Sec.~\ref{sec:iou-metric}, candidate pairs are selected according to the maximum overlap criterion
\begin{equation}
(i^*, j^*) = \arg\max_{i \neq j} I_{ij}.
\end{equation}
This rule selects the pair of slots that exhibits the strongest spatial competition.
Starting from the current slot set, the merge operator is applied to the selected pair $(i^*, j^*)$. After each merge, the attention matrix and slot representations are updated and the pairwise overlap scores are recomputed. The procedure is repeated while the maximum overlap among all slot pairs exceeds a predefined threshold. Since merges are performed sequentially, the procedure is not transitive: if slot $j$ overlaps with both $i$ and $k$, merging $j$ into $i$ does not necessarily imply that the resulting slot will also merge with $k$.

Let $I_{\max} = \max_{i \neq j} I_{ij}$ denote the maximum pairwise overlap in the current slot set. Merging terminates when
\(
I_{\max} \leq \tau ,
\)
where $\tau$ is a fixed threshold controlling the merging procedure. The threshold is estimated from the empirical distribution of pairwise overlap scores observed during training. Specifically, overlap values are aggregated over batches to form a histogram, and the triangle thresholding method~\cite{zack1977automatic} is applied to determine a separation point between low-overlap and high-overlap slot pairs.

\begin{figure}
    \centering
    \includegraphics[width=0.9\linewidth]{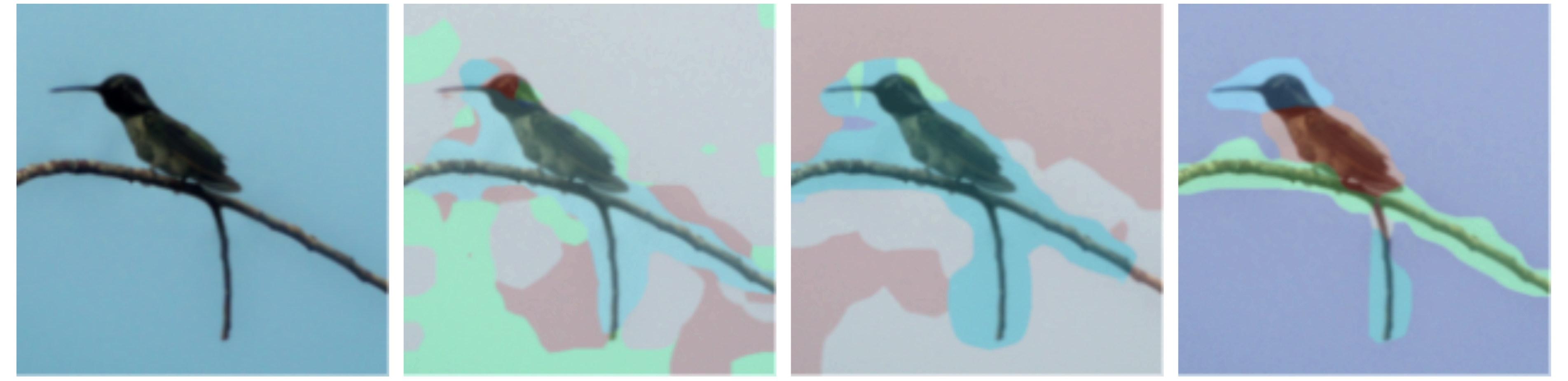}
    \caption{Evolution of slot representations during training (left to right). Slot assignments transition from fragmented and unstable to progressively specialized and object-aligned. After stabilization (right), slot merging is enabled to resolve overlap between slots.}
    \label{fig:convergence}
\end{figure}

\noindent\textbf{Training Schedule.}
To avoid interference with early slot formation, merging is activated only after the slot representations have stabilized (Fig.~\ref{fig:convergence}). In practice, the merge policy is enabled after an initial training phase and subsequently applied during the remaining optimization steps. Alg.~\ref{alg:slot_merging} summarizes the resulting merge procedure. The algorithm maintains a binary mask $M$ indicating active slots for each batch element, while $\mathcal{K}$ denotes the set of currently active slot indices.

\begin{algorithm}[t]
\caption{Slot Merging}
\label{alg:slot_merging}
\scalebox{0.8}{
\begin{minipage}{.95\linewidth}
\begin{algorithmic}[1]
\Require \begin{tabular}[t]{@{}l@{}}
$\text{slots } S^{(T)} \in \mathbb{R}^{B \times K \times D_{\text{slots}}}$,\\
$\text{attention maps } A \in \mathbb{R}^{B \times N \times K}$, threshold $\tau$
\end{tabular}

\State $S \gets S^{(T)}$ ; $M \gets \mathbf{1}^{B \times K}$

\For{$b=1,\dots,B$ \textbf{in parallel}}
\State $\mathcal{K} \gets \{ i \in [K] : M_{b,i}=1 \}$
\Comment{$\mathcal{K}$ denotes active slots}
    \While{$|\mathcal{K}|> 1$}
        \State compute $\{ I^{(b)}_{ij} \}_{i<j,\, i,j\in\mathcal{K}}$ via Eq.~\ref{eq:soft_iou}
        \State $(i^\ast,j^\ast) \gets
      \arg\max_{i<j,\, i,j\in\mathcal{K}} I^{(b)}_{ij}$
        \If{$I^{(b)}_{i^\ast, j^\ast} \le \tau$}
            \State \textbf{break}
        \EndIf
        \State compute masses $\alpha_{b,i^\ast}, \alpha_{b,j^\ast}$ via Eq.~\ref{eq:mass}
        \State compute weights $w_{b,i^\ast}, w_{b,j^\ast}$ from $\alpha_{b,i^\ast}, \alpha_{b,j^\ast}$ via Eq.~\ref{eq:convweights}
        \State $S_{b,i^\ast} \gets w_{b,i^\ast} S_{b,i^\ast} + w_{b,j^\ast} S_{b,j^\ast}$ via Eq.~\ref{eq:barymerge}
        \State $A_{b,:,i^\ast} \gets A_{b,:,i^\ast} + A_{b,:,j^\ast}$ via Eq.~\ref{eq:attnmapsupdate}
        \State $M_{b,j^\ast} \gets 0$ %M_{b,i^\ast} \gets 1,\; 
        \State $\mathcal{K} \gets \mathcal{K} \setminus \{j^\ast\}$
    \EndWhile
\EndFor

\State \Return $S, M$
\end{algorithmic}
\end{minipage}}
\end{algorithm}

\begin{table}[t]
\caption{Comparison between inference-only and training-integrated slot merging on VOC and MOVi-C, evaluated with mBO (instance and class). Results are averaged over 3 seeds. Unless otherwise specified, \emph{Ours} denotes DINOSAUR with the proposed slot merging mechanism.}
\label{tab:merge_on_inference}
\centering
\setlength{\tabcolsep}{4pt}
\resizebox{\linewidth}{!}{%
\begin{tabular}{l @{\hskip 0.8em} ccc @{\hskip 0.8em} cc}
\toprule
& \multicolumn{3}{c}{VOC} & \multicolumn{2}{c}{MOVi-C} \\
\cmidrule(lr){2-4} \cmidrule(lr){5-6}
& mBO$^i$ & mBO$^c$ & mIoU & mBO$^i$ & mIoU \\
\midrule
Baseline
& 44.0{\tiny$\pm$1.9} & 51.2{\tiny$\pm$1.9} & -- & 42.4 & -- \\
Inference
& 46.01{\tiny$\pm$0.7} & 52.68{\tiny$\pm$0.65} & 44.51{\tiny$\pm$0.55}
& 44.53{\tiny$\pm$1.7} & 43.81{\tiny$\pm$1.6} \\
\rowcolor{oursGreen}
Training
& \bf{47.83{\tiny$\pm$0.17}} & \bf{55.34{\tiny$\pm$0.27}} & \bf{46.01{\tiny$\pm$0.18}}
& \bf{48.35{\tiny$\pm$1.32}} & \bf{47.43{\tiny$\pm$1.26}} \\
\bottomrule
\end{tabular}%
}
\end{table}

\subsection{Computational Complexity}
\label{sec:complexity}
Slot Attention has time complexity $\mathcal{O}(T D N K)$\cite{locatello2020object}. Let $N$ denote the number of input patches and $K$ the initial number of slots, and let $m \le K-1$ denote the number of merges performed. A naive implementation recomputes all pairwise overlaps after each merge. At a stage with $r$ active slots there are $\mathcal{O}(r^2)$ slot pairs and each overlap requires a linear pass over $N$ patches, giving a cost $\mathcal{O}(N r^2)$ per step. Summing across merge steps yields $\sum_{r=K-m+1}^{K} \mathcal{O}(N r^2)$, which in the worst case $m=K-1$ becomes $\mathcal{O}(N K^3)$. In practice we implement merging with incremental updates. All pairwise soft intersections $\sum_{n}A_{n,i}A_{n,j}$ and slot masses $\sum_{n}A_{n,i}$ are precomputed once in $\mathcal{O}(NK^2)$. After contracting slot $j$ into $i$ via Eq.~\ref{eq:attnmapsupdate}, only overlaps incident to slot $i$ must be updated, which requires $\mathcal{O}(N K)$ work per merge. This yields an overall complexity of $\mathcal{O}(N K^2)$ per image. Consequently, the full pipeline scales as $\mathcal{O}(T D N K + N K^2)$ in practice. Since the number of slots is typically small, the additional merging cost remains minor compared to the Slot Attention updates.
\section{Experiments}
\label{sec:experiments}

\begin{figure}
    \centering
    \includegraphics[width=0.99\linewidth]{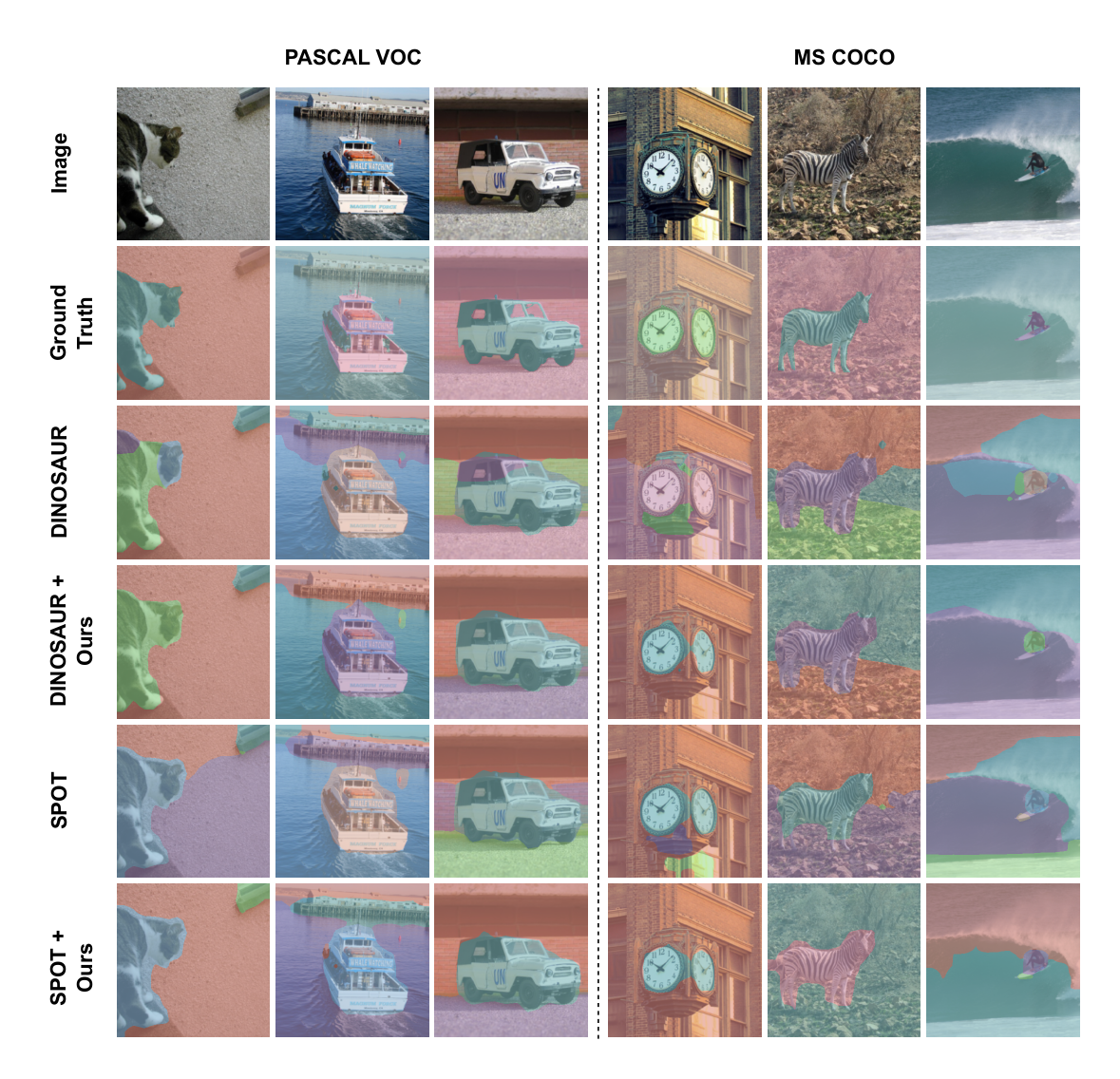}
    \caption{Qualitative results on COCO and PASCAL VOC.}
    \label{fig:images}
\end{figure}

\subsection{Setup}

\noindent\textbf{Datasets.}
We evaluate object discovery on four public benchmarks covering real-world and synthetic scenes. PASCAL VOC 2012~\cite{everingham2012pascal} and MS COCO 2017~\cite{lin2014microsoft} include natural scenes. MOVi-C and MOVi-E~\cite{greff2022kubric} are synthetic datasets generated from approximately 1,000 scanned 3D assets, with MOVi-E containing substantially more objects per scene than MOVi-C. Following~\cite{seitzer2022bridging}, we sample individual frames and treat them as independent images. For compositional generation and slot-level scene editing, we use CLEVRTex~\cite{karazija2clevrtex}, a texture-rich synthetic benchmark. Further dataset and preprocessing details are provided in the supplementary material.

\noindent\textbf{Evaluation Metrics.}
Following standard practice in object-centric learning, segmentation quality is evaluated using mean Best Overlap (mBO) and mean Intersection over Union (mIoU). mBO~\cite{pont2016multiscale} measures how well each ground-truth mask is recovered by selecting the predicted mask with the highest overlap. Both instance-level ($\mathrm{mBO}^i$) and class-level ($\mathrm{mBO}^c$) variants are reported. mIoU computes one-to-one assignments between predicted and ground-truth masks using the Hungarian matching algorithm, and averages the IoU over the matched pairs. Consistent with recent literature~\cite{tian2025pay,Kakogeorgiou_2024_CVPR}, Foreground Adjusted Rand Index (FG-ARI) is not reported, as it is known to favor under-segmentation and can therefore yield misleading conclusions (e.g., in MOVI-E we double FG-ARI from 32.24 to 64.11). For set prediction, we report top-1 and top-5 class accuracy together with the $R^2$ scores of the predicted object-center coordinates. For object recognition, we report top-1, and top-3 class accuracy, and bounding-box IoU. Compositional generation is evaluated using Fréchet Inception Distance (FID), while slot-level editing is assessed qualitatively.

\noindent\textbf{Implementation Details.}
The proposed slot merging mechanism is integrated into both DINOSAUR~\cite{seitzer2022bridging} and SPOT~\cite{Kakogeorgiou_2024_CVPR} architectures. Both models employ a ViT-B/16 encoder initialized with DINO~\cite{dosovitskiy2020image,caron2021emerging}. The number of initial slots is $K=7$, $8$, $12$, and $25$ for VOC, COCO, MOVi-C, and MOVi-E, respectively. DINOSAUR is trained for 100 epochs on VOC and COCO and 65 epochs on MOVi-C and MOVi-E. SPOT follows a two-stage teacher-student schedule: the teacher and student are trained for $560+560$ epochs on VOC, $50+50$ epochs on COCO, and $65+30$ epochs on both MOVi datasets. Merging is applied after the third Slot Attention iteration and enabled only after the slot representations stabilize (see Fig.~\ref{fig:convergence}). The merging threshold is estimated separately for each dataset from the distribution of pairwise overlap scores. For SPOT, merging is applied to the teacher during training and distillation, while the student uses merging only at inference to preserve a stable distillation target. Further implementation and training details are provided in the supplementary material.

\subsection{Analysis}

\noindent\textbf{Effect of Merging During Training.}
Tab.~\ref{tab:merge_on_inference} compares three configurations: the vanilla baseline (DINOSAUR), a variant with merging applied only at inference, and the full model with merging incorporated during training. Applying merging only at inference improves performance over the baseline. The ablation suggests that the proposed strategy is effective without modifying the training procedure or requiring additional fine-tuning. Integrating merging during training further improves all metrics across both datasets. On VOC, our method reaches $55.34$ $\mathrm{mBO}^c$, outperforming both the baseline and the inference-only variant. Deactivating slots without merging results in $51.15$ $\mathrm{mBO}^c$ on VOC, close to the baseline's $51.2$. This comparison suggests that combining slot representations contributes to the improvement beyond reducing the number of active slots. These results support the benefit of merging during training beyond its use as a post-processing step. Fig.~\ref{fig:images} visualizes the object-segmentation results of our method in comparison with DINOSAUR and SPOT.

\begin{table}[t]
\caption{Ablation on gradient detachment in the barycentric slot merging operator during training on VOC and MOVi-C. Results are averaged across 3 seeds.}
\label{tab:detach_ablation}
\centering
\setlength{\tabcolsep}{4pt}
\resizebox{\linewidth}{!}{%
\begin{tabular}{l @{\hskip 0.6em} ccc @{\hskip 0.6em} cc}
\toprule
& \multicolumn{3}{c}{VOC} & \multicolumn{2}{c}{MOVi-C} \\
\cmidrule(lr){2-4} \cmidrule(lr){5-6}
& mBO$^i$ & mBO$^c$ & mIoU & mBO$^i$ & mIoU \\
\midrule
w/ Detach
& 45.75{\tiny$\pm$0.28} & 53.13{\tiny$\pm$0.42} & 43.97{\tiny$\pm$0.22}
& 45.39{\tiny$\pm$2.95} & 44.46{\tiny$\pm$2.9} \\
\rowcolor{oursGreen}
w/o Detach
& \bf{47.83{\tiny$\pm$0.17}} & \bf{55.34{\tiny$\pm$0.27}} & \bf{46.01{\tiny$\pm$0.18}}
& \bf{48.35{\tiny$\pm$1.32}} & \bf{47.43{\tiny$\pm$1.26}} \\
\bottomrule
\end{tabular}%
}
\end{table}

\noindent\textbf{Effect of Differentiable Merging.} To verify that the observed gains arise from gradient propagation rather than from the merging rule itself, we conduct an ablation in which gradients are detached during the merging step (Tab.~\ref{tab:detach_ablation}). In this variant, merged slots weights are removed from the computational graph, preventing gradients from flowing back to the respective slots. Detaching gradients consistently degrades performance, which indicates that the improvement cannot be attributed to the forward merging operation alone. Allowing gradients to propagate through the merging layer ensures that slot representations are optimized jointly with the merging operation during training.

\begin{table}[b]
\caption{Comparison between uniform averaging and barycentric fusion during slot merging. Results are averaged over 3 seeds.}
\label{tab:barycentric_fusion}
\centering
{
\begin{tabular}{l p{0.6em} ccc}
\toprule
& & mBO$^i$ & mBO$^c$ & mIoU \\
\midrule
Uniform Avg.
& & 45.92{\tiny$\pm$0.23} & 53.38{\tiny$\pm$0.39} & 44.07{\tiny$\pm$0.24} \\

\rowcolor{oursGreen}
Barycentric
& & \bf{47.83{\tiny$\pm$0.17}} & \bf{55.34{\tiny$\pm$0.27}} & \bf{46.01{\tiny$\pm$0.18}} \\
\bottomrule
\end{tabular}
}
\end{table}

\noindent\textbf{Effect of Barycentric Fusion.}
To evaluate the mass-weighted slot update, we compare the proposed barycentric fusion in Eq.~\ref{eq:barymerge} with uniform averaging. It assigns equal importance to both slots, while barycentric fusion weights each representation according to its attention mass. As shown in Tab.~\ref{tab:barycentric_fusion}, the proposed update consistently improves performance, demonstrating the importance of accounting for the spatial contribution of each slot during merging.

\begin{figure}
    \centering
    \includegraphics[width=0.98\linewidth]{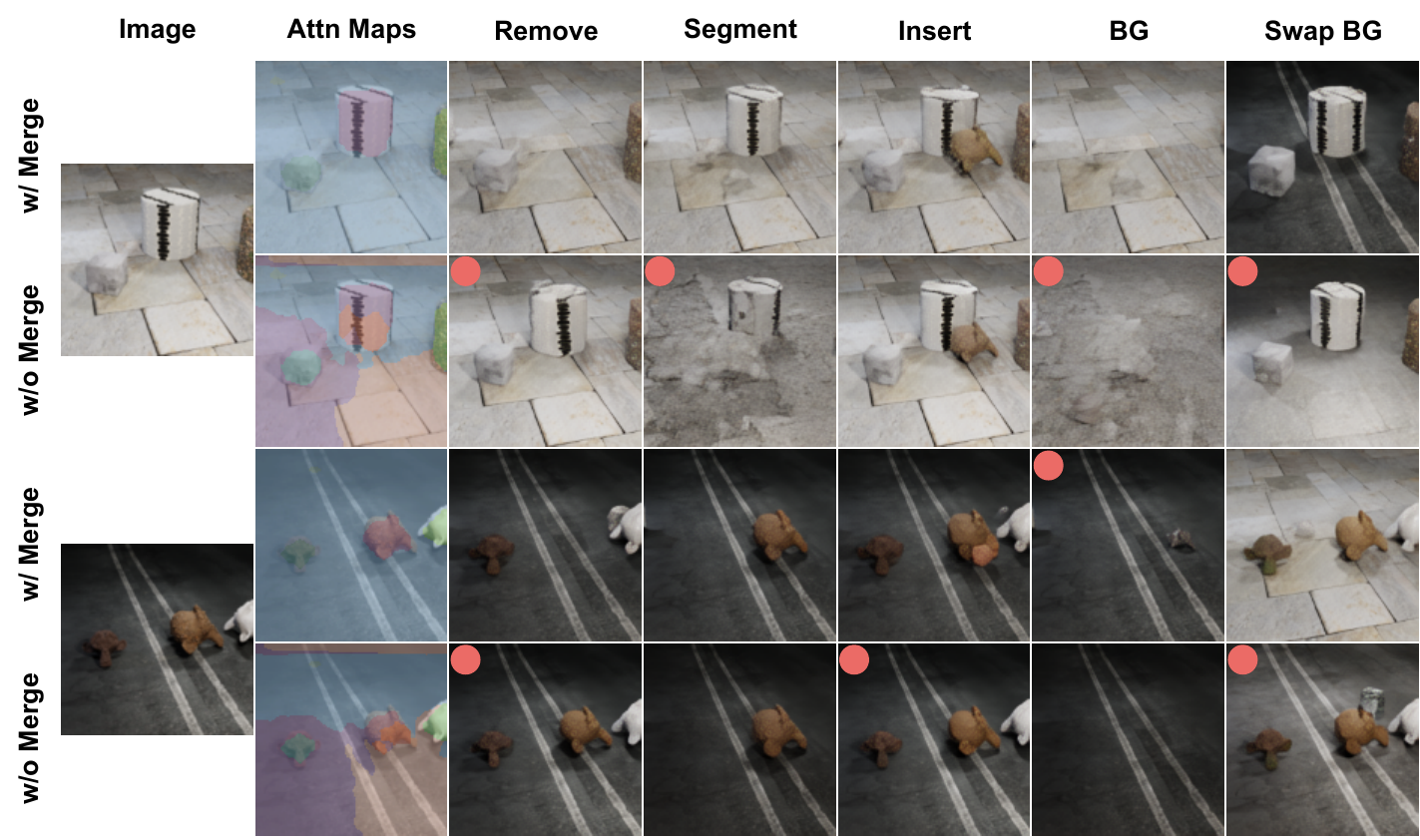}
    \caption{Slot-level compositional editing in SlotDiffusion. For each scene, the left image shows the input, while the grid compares vanilla slots and our merged slots. Columns show predicted masks, object removal, object segmentation, object insertion, background-only generation, and background swap. \textcolor{failred}{Red dots} mark visible editing failures, such as fragmented object edits, residual foreground leakage, or inconsistent background transfer. Merging yields more object-complete manipulations and cleaner background edits.
}
    \label{fig:slotdiff}
\end{figure}

\begin{table}[b]
\caption{Effect of slot merging on compositional generation and object discovery in SlotDiffusion. FID is computed for compositional generation; segmentation is evaluated with mIoU and mBO$^i$.}
\label{tab:merge_fid_segmentation}
\centering
\setlength{\tabcolsep}{5pt}
{%
\begin{tabular}{l @{\hskip 1.0em} c @{\hskip 1.0em} cc}
\toprule
& \multicolumn{1}{c}{Generation} & \multicolumn{2}{c}{Object Discovery} \\
\cmidrule(lr){2-2} \cmidrule(lr){3-4}
Method
& FID $\downarrow$
& mIoU 
& mBO$^i$  \\
\midrule
SlotDiffusion~\cite{wu2023slotdiffusion}
& 35.26
& 57.20
& \bf{61.50} \\
\rowcolor{oursGreen}
SlotDiffusion $+$ Ours 
& \bf{29.74}
& \bf{64.47}
& 60.50 \\
\bottomrule
\end{tabular}%
}
\end{table}

\begin{table*}[ht]
\centering
\caption{Comparison with object-centric methods on VOC, COCO, MOVi-C, and MOVi-E. Results for our approach are reported as mean $\pm$ standard deviation over 3 seeds, using a ViT-B/16 backbone and $224 \times 224$ input resolution. $\dag$ indicates results taken from~\cite{Kakogeorgiou_2024_CVPR}. Otherwise, results are from the original papers. The best and second-best results are shown in \textbf{bold} and \underline{underlined}, respectively.}
\label{tab:main_results}

\setlength{\tabcolsep}{4pt}
\resizebox{\textwidth}{!}{%
\begin{tabular}{l @{\hskip 0.8em} ccc @{\hskip 0.8em} ccc @{\hskip 0.8em} cc @{\hskip 0.8em} cc}
\toprule
& \multicolumn{3}{c}{VOC}
& \multicolumn{3}{c}{COCO}
& \multicolumn{2}{c}{MOVi-C}
& \multicolumn{2}{c}{MOVi-E} \\
\cmidrule(lr){2-4}
\cmidrule(lr){5-7}
\cmidrule(lr){8-9}
\cmidrule(lr){10-11}
Method
& mBO$^i$ & mBO$^c$ & mIoU
& mBO$^i$ & mBO$^c$ & mIoU
& mBO$^i$ & mIoU
& mBO$^i$ & mIoU \\
\midrule

SA \cite{locatello2020object}$^{\dag}$ &
24.6 & 24.9 & -- &
17.2 & 19.2 & -- &
26.2{\tiny$\pm$1.0} & -- &
24.0{\tiny$\pm$1.2} & -- \\

SLATE \cite{singh2021illiterate}$^{\dag}$ &
35.9 & 41.5 & -- &
29.1 & 33.6 & -- &
39.4{\tiny$\pm$0.8} & 37.8{\tiny$\pm$0.7} &
30.2{\tiny$\pm$1.7} & 28.6{\tiny$\pm$1.7} \\

DINOSAUR-MLP \cite{seitzer2022bridging}$^{\dag}$ &
39.5{\tiny$\pm$0.1} & 40.9{\tiny$\pm$0.1} & -- &
27.7{\tiny$\pm$0.2} & 30.9{\tiny$\pm$0.2} & -- &
39.1{\tiny$\pm$0.2} & -- &
35.5{\tiny$\pm$0.2} & -- \\

DINOSAUR \cite{seitzer2022bridging}$^{\dag}$ &
44.0{\tiny$\pm$1.9} & 51.2{\tiny$\pm$1.9} & -- &
32.3{\tiny$\pm$0.4} & 38.8{\tiny$\pm$0.4} & -- &
42.4 & -- &
-- & -- \\

SPOT~\cite{Kakogeorgiou_2024_CVPR} &
\underline{48.3{\tiny$\pm$0.4}} &
\underline{55.6{\tiny$\pm$0.4}} &
\underline{46.8{\tiny$\pm$0.4}} &
\textbf{35.0{\tiny$\pm$0.1}} &
\underline{44.7{\tiny$\pm$0.3}} &
\textbf{33.0{\tiny$\pm$0.1}} &
47.3{\tiny$\pm$1.2} & 46.7{\tiny$\pm$1.3} &
\underline{40.1{\tiny$\pm$1.2}} &
\underline{39.3{\tiny$\pm$1.2}} \\

\midrule

\rowcolor{oursGreen}
\textbf{DINOSAUR + Ours} &
47.83{\tiny$\pm$0.17} &
55.34{\tiny$\pm$0.27} &
46.01{\tiny$\pm$0.18} &
32.88{\tiny$\pm$0.40} &
42.78{\tiny$\pm$0.22} &
30.75{\tiny$\pm$0.55} &
\underline{48.35{\tiny$\pm$1.32}} &
\underline{47.43{\tiny$\pm$1.26}} &
39.56{\tiny$\pm$0.55} &
37.75{\tiny$\pm$0.55} \\

\rowcolor{oursGreen}
\textbf{SPOT + Ours} &
\textbf{49.44{\tiny$\pm$0.63}} &
\textbf{57.35{\tiny$\pm$0.66}} &
\textbf{47.62{\tiny$\pm$0.62}} &
\underline{34.77{\tiny$\pm$0.41}} &
\textbf{45.18{\tiny$\pm$0.35}} &
\underline{32.54{\tiny$\pm$0.46}} &
\textbf{49.50{\tiny$\pm$0.52}} &
\textbf{48.17{\tiny$\pm$0.83}} &
\textbf{41.39{\tiny$\pm$0.42}} &
\textbf{39.94{\tiny$\pm$0.45}} \\

\bottomrule
\end{tabular}%
}
\end{table*}

\subsection{Comparison with Object-Centric Methods}
Tab.~\ref{tab:main_results} compares slot merging with the underlying baselines and prior object-centric learning approaches. Integrating merging into DINOSAUR, consistently improves performance across all four datasets, with gains of $4.14$ points in $\mathrm{mBO}^c$ and $3.83$ points in $\mathrm{mBO}^i$ on VOC. On COCO, $\mathrm{mBO}^c$ increases from $38.8$ to $42.78$, while on MOVi-C, $\mathrm{mBO}^i$ increases from $42.4$ to $48.35$. Integrating into SPOT, increases $\mathrm{mBO}^c$ from $55.6$ to $57.35$ on VOC and $\mathrm{mBO}^i$ from $47.3$ to $49.50$ on MOVi-C, showing that the gains extend across different object-centric learning architectures.

\begin{table}[t]
\centering
\caption{Comparison with adaptive object-centric methods on MOVi-C and MOVi-E.
Results for our approach are reported as mean $\pm$ standard deviation over 3 seeds,
using a ViT-B/16 backbone and $224 \times 224$ input resolution.
Results are taken from the methods' original papers. Best and second-best results are shown in bold
and underlined, respectively.}
\label{tab:more_results}

\setlength{\tabcolsep}{4pt}
\resizebox{\columnwidth}{!}{%
\begin{tabular}{l @{\hskip 0.8em} cc @{\hskip 0.8em} cc}
\toprule
& \multicolumn{2}{c}{MOVi-C}
& \multicolumn{2}{c}{MOVi-E} \\
\cmidrule(lr){2-3}
\cmidrule(lr){4-5}
Method
& mBO$^i$ & mIoU
& mBO$^i$ & mIoU \\
\midrule

AdaSlot~\cite{fan2024adaptive} &
35.6 & -- &
29.8 & -- \\

MetaSlot-MLP~\cite{liu2025metaslot} &
35.0 & -- &
-- & -- \\

MetaSlot-Tfd~\cite{liu2025metaslot} &
-- & -- &
-- & -- \\

MetaSlot-Dfz~\cite{liu2025metaslot} &
-- & -- &
-- & -- \\

DINOSAUR + FS + RC~\cite{tian2025pay} &
46.0 & 44.8 &
-- & -- \\

SPOT + FS + RC~\cite{tian2025pay} &
\underline{49.0} & \underline{47.8} &
-- & -- \\

QASA~\cite{ouyang2026qasa} &
46.9 & 46.1 &
39.1 & \underline{37.8} \\

QASA-MLP~\cite{ouyang2026qasa} &
42.2 & 41.1 &
34.6 & 33.0 \\

\midrule

\rowcolor{oursGreen}
\textbf{DINOSAUR + Ours} &
48.35{\tiny$\pm$1.32} &
47.43{\tiny$\pm$1.26} &
\underline{39.56{\tiny$\pm$0.55}} &
37.75{\tiny$\pm$0.55} \\

\rowcolor{oursGreen}
\textbf{SPOT + Ours} &
\textbf{49.50{\tiny$\pm$0.52}} &
\textbf{48.17{\tiny$\pm$0.83}} &
\textbf{41.39{\tiny$\pm$0.42}} &
\textbf{39.94{\tiny$\pm$0.45}} \\

\bottomrule
\end{tabular}%
}
\end{table}

Beyond these baseline improvements, Tab.~\ref{tab:more_results} compares our method with approaches that adapt the scene decomposition through slot selection in AdaSlot~\cite{fan2024adaptive} and QASA~\cite{ouyang2026qasa}, prototype-based deduplication in MetaSlot~\cite{liu2025metaslot}, and region combination in FS+RC~\cite{tian2025pay}. The variable object counts and larger initial slot budgets of MOVi-C and MOVi-E make the treatment of redundant slots particularly important. Our method achieves competitive results across both datasets, outperforming the compared approaches on several object discovery metrics.

\subsection{Applications}
\label{sec:applications}

We evaluate whether slot merging improves the usefulness of learned slots beyond object discovery and segmentation, through set prediction and object recognition on PASCAL VOC, and compositional generation and slot-level scene editing on CLEVRTex. For the two supervised probes, the encoder and Slot Attention module remain frozen, and only lightweight slot-wise prediction heads are trained. Compositional generation and slot-level scene editing use a pretrained SlotDiffusion model. Further details on the downstream tasks and evaluation protocols are provided in the supplementary material.

\begin{table}[t]
\centering\footnotesize
\setlength{\tabcolsep}{4pt}
\caption{Results in \textbf{(Top)} Object recognition and
\textbf{(Bottom)} Set prediction. All metrics are reported $\times100$. Results are reported as mean $\pm$ standard deviation over 3 seeds.}
\label{tab:aux}
\begin{tabular}{l cccc}
\toprule
Object Recognition & top-1 & top-3 & bbox IoU  \\
\midrule
DINOSAUR (VOC) & 80.5{\tiny$\pm$0.2}  & 94.9{\tiny$\pm$0.3}  &  48.8{\tiny$\pm$0.1} \\
\rowcolor{oursGreen}\quad $+$ Ours & \bf{83.0{\tiny$\pm$1.3}} & \bf{96.2{\tiny$\pm$0.4}}  & \bf{50.4{\tiny$\pm$1.0}}  \\
\midrule
Set Prediction & top-1  & top-5 & $R^2(x)$ & $R^2(y)$ \\
\midrule
DINOSAUR (VOC) & 74.3{\tiny$\pm$0.9} & 94.1{\tiny$\pm$0.1} & 24.5{\tiny$\pm$1.4} & 23.9{\tiny$\pm$1.0} \\
\rowcolor{oursGreen}\quad $+$ Ours  & \bf{77.0{\tiny$\pm$0.8}} & \bf{96.4{\tiny$\pm$0.1}} & \bf{26.6{\tiny$\pm$0.9}} & \bf{25.8{\tiny$\pm$1.3}}  \\
\bottomrule
\end{tabular}
\end{table}
\begin{figure*}
    \centering
    \includegraphics[width=0.99\linewidth]{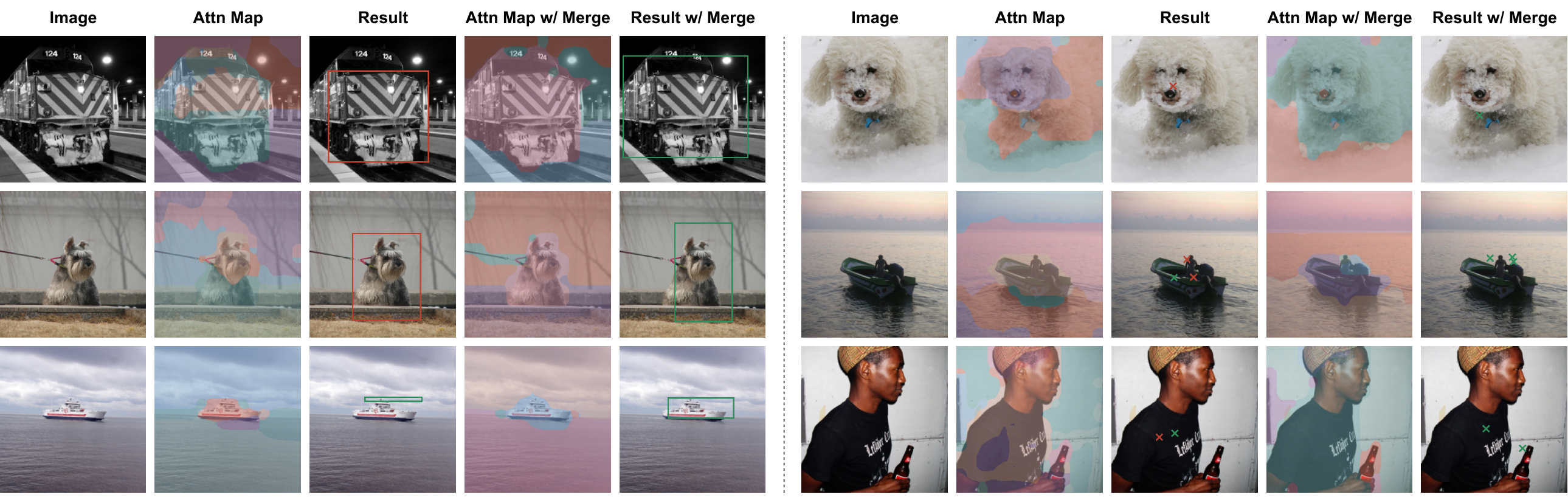}
    \caption{Qualitative results for object recognition and set prediction on PASCAL VOC. Each example shows the input image, slot attention, and downstream prediction produced by DINOSAUR with and without slot merging. In object recognition (left group), boxes denote predicted object extents. In set prediction (right group), crosses denote predicted object centers. Green boxes or crosses indicate \textcolor{ForestGreen}{correct class predictions}, while red shows \textcolor{BrickRed}{wrong predictions}. Spatial placement measures localization quality. Slot merging produces more complete object support and preserves more relevant context, leading to improved localization and classification.}
    \label{fig:sa_apps}
\end{figure*}

\noindent\textbf{Set Prediction.} As shown in Tab.~\ref{tab:aux}, slot merging improves set prediction metrics, for example Top-1 accuracy increases from $74.3$ to $77.0$. The qualitative results in the right group of Fig.~\ref{fig:sa_apps} show that merging results in more coherent object slots. In the second row, the baseline associates both people and part of the boat with the same slot, whereas slot merging separates the two people and localizes both correctly. In the third row, the person and bottle are represented by distinct slots and are both correctly localized and classified. 

\noindent\textbf{Object Recognition.} In Fig.~\ref{fig:sa_apps} the left group illustrates the effect of merging in object recognition. For the dog example (second row), the baseline assigns $44\%$ probability to sheep and only $18\%$ to dog, whereas the merged representation assigns $67\%$ to dog and $17\%$ to sheep. The merged attention map covers the object more coherently and retains relevant context, providing the prediction head with a cleaner representation for both classification and localization.

\noindent\textbf{Compositional Generation.} We apply slot merging to a pretrained SlotDiffusion model~\cite{wu2023slotdiffusion} without further optimization. As shown in Tab.~\ref{tab:merge_fid_segmentation}, merging reduces compositional-generation FID from $35.26$ to $29.74$. It also increases mIoU from $57.20$ to $64.47$, a gain of $7.27$ points. These results show a substantial improvement in compositional generation and mIoU, with a small trade-off in instance-level overlap. We further apply slot merging to GLASS~\cite{singh2025glass}, another diffusion-based architecture, improving object discovery on COCO without additional training (see supplementary material).

\noindent\textbf{Slot-Level Scene Editing.} Fig.~\ref{fig:slotdiff} shows that more coherent slots also enable more controlled editing. In the "Remove" examples, the SlotDiffusion baseline cannot fully remove the selected object because its content is distributed across several slots, whereas slot merging isolates and removes the complete object. The same difference appears in object segmentation and background isolation, where merged slots preserve the full object more consistently. During background swapping, the baseline retains elements from both backgrounds because their information is spread across multiple slots, whereas slot merging isolates and transfers the background more accurately. These results connect improved slot factorization with more reliable removal, segmentation, insertion, and background manipulation.

\subsection{Limitations}
The proposed merging mechanism is intentionally simple and relies on pairwise overlap statistics to regulate slot consolidation. While this design keeps the procedure efficient, it does not explicitly model higher-order relations among multiple slots, since merging decisions are based on the overlap between individual slot pairs. Extending the policy to incorporate group-level similarity or set-based distance measures could provide additional context when resolving complex interactions between objects. More broadly, the current stopping rule is derived from the empirical distribution of overlap scores and remains fixed during training, without adapting to changes in the slot representations. Developing more interpretable or adaptive stopping criteria that account for scene composition or global structure represents an interesting direction for future work.
\section{Conclusion}
\label{sec:conclusion}

Slot Attention specifies how slots compose an image, but it does not define operations on the slot set itself. As a result, representational fragmentation may happen, where multiple slots bind to the same entity and spend capacity resolving mutual competition. To address this, we introduce a merge operator that acts directly on the slot set. The operator identifies overlapping slots and replaces them with a single aggregated representation, converting competitive redundancy into cooperative refinement at the representation level. The proposed mechanism is simple, local, and leaves the underlying Slot Attention dynamics unchanged. Empirically, merging stabilizes slot factorization and improves object-level grouping across datasets, resulting in more useful representations. More broadly, our results highlight the importance of explicit structural operations on the slot set, complementing existing efforts that focus on improving feature binding or architectural design.
\section*{Acknowledgments}
This work was partially supported by the TEXTaiLES project under the European Union’s Horizon Europe research and innovation programme (Grant Agreement No. 101158328) and by project MIS 5154714 of the National Recovery and Resilience Plan Greece 2.0, funded by the European Union through NextGenerationEU. This work also received partial support through the Pharos AI Factory project from the European High-Performance Computing Joint Undertaking (EuroHPC JU) under Grant Agreement No. 101234269 and from the Greek Ministry of Digital Governance and Artificial Intelligence. The authors acknowledge the EuroHPC JU for granting access to the LEONARDO supercomputer at CINECA, Italy, through a EuroHPC Development Access call (project EHPC-DEV-2026D05-011).
\section*{LLM Usage Statement}
The authors acknowledge the usage of OpenAI's ChatGPT to correct grammar, to edit the related work and experimental descriptions, and additionally Anthropic's Claude to obtain reviewer style feedback on the manuscript. ChatGPT was also used to format tables, to write code to generate the 1D plots of attention scores in Fig.~\ref{fig:merging_process}, and also automate the visualization process.
{
    \small
    \bibliographystyle{ieeenat_fullname}
    \bibliography{main}

@String(CVPR= {IEEE Conf. Comput. Vis. Pattern Recog.})

@String(ICLR = {Int. Conf. Learn. Represent.})

@String(CVPR  = {CVPR})

@String(ICLR  = {ICLR})

@article{locatello2020object,
  title={Object-centric learning with slot attention},
  author={Locatello, Francesco and Weissenborn, Dirk and Unterthiner, Thomas and Mahendran, Aravindh and Heigold, Georg and Uszkoreit, Jakob and Dosovitskiy, Alexey and Kipf, Thomas},
  journal={Advances in neural information processing systems},
  volume={33},
  pages={11525--11538},
  year={2020}
}

@inproceedings{fan2024adaptive,
  title={Adaptive slot attention: Object discovery with dynamic slot number},
  author={Fan, Ke and Bai, Zechen and Xiao, Tianjun and He, Tong and Horn, Max and Fu, Yanwei and Locatello, Francesco and Zhang, Zheng},
  booktitle={Proceedings of the IEEE/CVF Conference on Computer Vision and Pattern Recognition},
  pages={23062--23071},
  year={2024}
}

@inproceedings{liu2025metaslot,
  title={MetaSlot: Break Through the Fixed Number of Slots in Object-Centric Learning},
  author={Liu, Hongjia and Zhao, Rongzhen and Chen, Haohan and Pajarinen, Joni},
  booktitle={The Thirty-ninth Annual Conference on Neural Information Processing Systems},
  year={2025}
}

@article{zack1977automatic,
  title={Automatic measurement of sister chromatid exchange frequency.},
  author={Zack, Gregory W and Rogers, William E and Latt, Samuel A},
  journal={Journal of Histochemistry \& Cytochemistry},
  volume={25},
  number={7},
  pages={741--753},
  year={1977},
  publisher={SAGE Publications Sage CA: Los Angeles, CA}
}

@article{burgess2019monet,
  title={Monet: Unsupervised scene decomposition and representation},
  author={Burgess, Christopher P and Matthey, Loic and Watters, Nicholas and Kabra, Rishabh and Higgins, Irina and Botvinick, Matt and Lerchner, Alexander},
  journal={arXiv preprint arXiv:1901.11390},
  year={2019}
}

@article{eslami2016attend,
  title={Attend, infer, repeat: Fast scene understanding with generative models},
  author={Eslami, SM and Heess, Nicolas and Weber, Theophane and Tassa, Yuval and Szepesvari, David and Hinton, Geoffrey E and others},
  journal={Advances in neural information processing systems},
  volume={29},
  year={2016}
}

@article{kosiorek2018sequential,
  title={Sequential attend, infer, repeat: Generative modelling of moving objects},
  author={Kosiorek, Adam and Kim, Hyunjik and Teh, Yee Whye and Posner, Ingmar},
  journal={Advances in Neural Information Processing Systems},
  volume={31},
  year={2018}
}

@inproceedings{greff2019multi,
  title={Multi-object representation learning with iterative variational inference},
  author={Greff, Klaus and Kaufman, Rapha{\"e}l Lopez and Kabra, Rishabh and Watters, Nick and Burgess, Christopher and Zoran, Daniel and Matthey, Loic and Botvinick, Matthew and Lerchner, Alexander},
  booktitle={International conference on machine learning},
  pages={2424--2433},
  year={2019},
  organization={PMLR}
}

@inproceedings{engelcke2020genesis,
  title={{GENESIS: Generative Scene Inference and Sampling with Object-Centric Latent Representations}},
  author={Engelcke, Martin and Kosiorek, Adam R and Parker Jones, Oiwi and Posner, Ingmar},
  booktitle={International Conference on Learning Representations (ICLR)},
  year={2020}
}

@article{aydemir2023self,
  title={Self-supervised object-centric learning for videos},
  author={Aydemir, G{\"o}rkay and Xie, Weidi and Guney, Fatma},
  journal={Advances in Neural Information Processing Systems},
  volume={36},
  pages={32879--32899},
  year={2023}
}

@inproceedings{seitzer2022bridging,
  title={Bridging the gap to real-world object-centric learning},
  author={Seitzer, M and Horn, M and Zadaianchuk, A and Zietlow, D and Xiao, T and Simon-Gabriel, C and He, T and Zhang, Z and Sch{\"o}lkopf, B and Brox, Thomas and others},
  booktitle={International Conference on Learning Representations (ICLR)},
  year={2023}
}

@inproceedings{singh2021illiterate,
  title={Illiterate DALL-E Learns to Compose},
  author={Singh, Gautam and Deng, Fei and Ahn, Sungjin},
  booktitle={International Conference on Learning Representations},
  year = {2022}
}

@inproceedings{jiang2023object,
  title = {Object-Centric Slot Diffusion},
  author = {Jiang, Jindong and Deng, Fei and Singh, Gautam and Ahn, Sungjin},
  booktitle = {Advances in Neural Information Processing Systems},
  volume = {36},
  pages = {8563--8601},
  url = {https://proceedings.neurips.cc/paper_files/paper/2023/file/1b3ceb8a495a63ced4a48f8429ccdcd8-Paper-Conference.pdf},
  year = {2023}
}

@InProceedings{Kakogeorgiou_2024_CVPR,
    author    = {Kakogeorgiou, Ioannis and Gidaris, Spyros and Karantzalos, Konstantinos and Komodakis, Nikos},
    title     = {SPOT: Self-Training with Patch-Order Permutation for Object-Centric Learning with Autoregressive Transformers},
    booktitle = {Proceedings of the IEEE/CVF Conference on Computer Vision and Pattern Recognition (CVPR)},
    month     = {June},
    year      = {2024},
    pages     = {22776-22786}
}

@article{zadaianchuk2023object,
  title={Object-centric learning for real-world videos by predicting temporal feature similarities},
  author={Zadaianchuk, Andrii and Seitzer, Maximilian and Martius, Georg},
  journal={Advances in Neural Information Processing Systems},
  volume={36},
  pages={61514--61545},
  year={2023}
}

@inproceedings{kipf2022conditional,
    author = {Kipf, Thomas
              and Elsayed, Gamaleldin F.
              and Mahendran, Aravindh
              and Stone, Austin
              and Sabour, Sara
              and Heigold, Georg
              and Jonschkowski, Rico
              and Dosovitskiy, Alexey
              and Greff, Klaus},
    title = {{Conditional Object-Centric Learning from Video}},
    booktitle = {International Conference on Learning Representations (ICLR)},
    year  = {2022}
}

@article{engelcke2021genesis,
  title={Genesis-v2: Inferring unordered object representations without iterative refinement},
  author={Engelcke, Martin and Parker Jones, Oiwi and Posner, Ingmar},
  journal={Advances in Neural Information Processing Systems},
  volume={34},
  pages={8085--8094},
  year={2021}
}

@inproceedings{everingham2012pascal,
  title={The pascal visual object classes challenge 2012 (voc2012)},
  author={Everingham, Mark and Van Gool, Luc and Williams, Christopher KI and Winn, John and Zisserman, Andrew},
  booktitle={Results},
  year={2012}
}

@inproceedings{lin2014microsoft,
  title={Microsoft coco: Common objects in context},
  author={Lin, Tsung-Yi and Maire, Michael and Belongie, Serge and Hays, James and Perona, Pietro and Ramanan, Deva and Doll{\'a}r, Piotr and Zitnick, C Lawrence},
  booktitle={European conference on computer vision},
  pages={740--755},
  year={2014},
  organization={Springer}
}

@inproceedings{greff2022kubric,
  title={Kubric: A scalable dataset generator},
  author={Greff, Klaus and Belletti, Francois and Beyer, Lucas and Doersch, Carl and Du, Yilun and Duckworth, Daniel and Fleet, David J and Gnanapragasam, Dan and Golemo, Florian and Herrmann, Charles and others},
  booktitle={Proceedings of the IEEE/CVF conference on computer vision and pattern recognition},
  pages={3749--3761},
  year={2022}
}

@inproceedings{tian2025pay,
  title={Pay attention to the foreground in object-centric learning},
  author={Tian, Pinzhuo and Yang, Shengjie and Yu, Hang and Kot, Alex},
  booktitle={Proceedings of the Computer Vision and Pattern Recognition Conference},
  pages={30281--30290},
  year={2025}
}

@inproceedings{caron2021emerging,
  title={Emerging properties in self-supervised vision transformers},
  author={Caron, Mathilde and Touvron, Hugo and Misra, Ishan and J{\'e}gou, Herv{\'e} and Mairal, Julien and Bojanowski, Piotr and Joulin, Armand},
  booktitle={Proceedings of the IEEE/CVF international conference on computer vision},
  pages={9650--9660},
  year={2021}
}

@article{dosovitskiy2020image,
  title={An image is worth 16x16 words: Transformers for image recognition at scale},
  author={Dosovitskiy, Alexey},
  journal={arXiv preprint arXiv:2010.11929},
  year={2020}
}

@article{kim2024bootstrapping,
  title={Bootstrapping top-down information for self-modulating slot attention},
  author={Kim, Dongwon and Kim, Seoyeon and Kwak, Suha},
  journal={Advances in Neural Information Processing Systems},
  volume={37},
  pages={103751--103773},
  year={2024}
}

@article{chi2025slot,
  title={Slot-MLLM: Object-Centric Visual Tokenization for Multimodal LLM},
  author={Chi, Donghwan and Kim, Hyomin and Oh, Yoonjin and Kim, Yongjin and Lee, Donghoon and Jo, Daejin and Kim, Jongmin and Baek, Junyeob and Ahn, Sungjin and Kim, Sungwoong},
  journal={arXiv preprint arXiv:2505.17726},
  year={2025}
}

@article{rubinstein2025we,
  title={Are we done with object-centric learning?},
  author={Rubinstein, Alexander and Prabhu, Ameya and Bethge, Matthias and Oh, Seong Joon},
  journal={arXiv preprint arXiv:2504.07092},
  year={2025}
}

@inproceedings{chapin2025object,
  title={Is an object-centric representation beneficial for robotic manipulation?},
  author={Chapin, Alexandre and Dellandrea, Emmanuel and Chen, Liming},
  booktitle={International Conference on Robotics, Computer Vision and Intelligent Systems},
  pages={510--523},
  year={2025},
  organization={Springer}
}

@inproceedings{he2022masked,
  title={Masked autoencoders are scalable vision learners},
  author={He, Kaiming and Chen, Xinlei and Xie, Saining and Li, Yanghao and Doll{\'a}r, Piotr and Girshick, Ross},
  booktitle={Proceedings of the IEEE/CVF conference on computer vision and pattern recognition},
  pages={16000--16009},
  year={2022}
}

@article{oquab2023dinov2,
  title={Dinov2: Learning robust visual features without supervision},
  author={Oquab, Maxime and Darcet, Timoth{\'e}e and Moutakanni, Th{\'e}o and Vo, Huy and Szafraniec, Marc and Khalidov, Vasil and Fernandez, Pierre and Haziza, Daniel and Massa, Francisco and El-Nouby, Alaaeldin and others},
  journal={arXiv preprint arXiv:2304.07193},
  year={2023}
}

@inproceedings{rahman2016optimizing,
  title={Optimizing intersection-over-union in deep neural networks for image segmentation},
  author={Rahman, Md Atiqur and Wang, Yang},
  booktitle={International symposium on visual computing},
  pages={234--244},
  year={2016},
  organization={Springer}
}

@article{pont2016multiscale,
  title={Multiscale combinatorial grouping for image segmentation and object proposal generation},
  author={Pont-Tuset, Jordi and Arbelaez, Pablo and Barron, Jonathan T and Marques, Ferran and Malik, Jitendra},
  journal={IEEE transactions on pattern analysis and machine intelligence},
  volume={39},
  number={1},
  pages={128--140},
  year={2016},
  publisher={IEEE}
}

@inproceedings{zhao2025slot,
  title={Slot attention with re-initialization and self-distillation},
  author={Zhao, Rongzhen and Zhao, Yi and Kannala, Juho and Pajarinen, Joni},
  booktitle={Proceedings of the 33rd ACM International Conference on Multimedia},
  pages={4185--4192},
  year={2025}
}

@inproceedings{kim2023shepherding,
  title={Shepherding slots to objects: Towards stable and robust object-centric learning},
  author={Kim, Jinwoo and Choi, Janghyuk and Choi, Ho-Jin and Kim, Seon Joo},
  booktitle={Proceedings of the IEEE/CVF Conference on Computer Vision and Pattern Recognition},
  pages={19198--19207},
  year={2023}
}

@article{lowecomplex,
  title={Complex-Valued Autoencoders for Object Discovery},
  author={L{\"o}we, Sindy and Lippe, Phillip and Rudolph, Maja and Welling, Max},
  journal={Transactions on Machine Learning Research},
  year = {2022}
}

@article{lowe2023rotating,
  title={Rotating features for object discovery},
  author={L{\"o}we, Sindy and Lippe, Phillip and Locatello, Francesco and Welling, Max},
  journal={Advances in Neural Information Processing Systems},
  volume={36},
  pages={59606--59635},
  year={2023}
}

@inproceedings{singh2025glass,
  title={Glass: Guided latent slot diffusion for object-centric learning},
  author={Singh, Krishnakant and Schaub-Meyer, Simone and Roth, Stefan},
  booktitle={Proceedings of the Computer Vision and Pattern Recognition Conference},
  pages={28673--28683},
  year={2025}
}

@article{wu2023slotdiffusion,
  title={Slotdiffusion: Object-centric generative modeling with diffusion models},
  author={Wu, Ziyi and Hu, Jingyu and Lu, Wuyue and Gilitschenski, Igor and Garg, Animesh},
  journal={Advances in Neural Information Processing Systems},
  volume={36},
  pages={50932--50958},
  year={2023}
}

@inproceedings{hariharan2011semantic,
  title={Semantic contours from inverse detectors},
  author={Hariharan, Bharath and Arbel{\'a}ez, Pablo and Bourdev, Lubomir and Maji, Subhransu and Malik, Jitendra},
  booktitle={2011 international conference on computer vision},
  pages={991--998},
  year={2011},
  organization={IEEE}
}

@inproceedings{zhao2025vector,
  title={Vector-quantized vision foundation models for object-centric learning},
  author={Zhao, Rongzhen and Wang, Vivienne Huiling and Kannala, Juho and Pajarinen, Joni},
  booktitle={Proceedings of the 33rd ACM International Conference on Multimedia},
  pages={5422--5430},
  year={2025}
}

@article{elsayed2022savi++,
  title={Savi++: Towards end-to-end object-centric learning from real-world videos},
  author={Elsayed, Gamaleldin and Mahendran, Aravindh and Van Steenkiste, Sjoerd and Greff, Klaus and Mozer, Michael C and Kipf, Thomas},
  journal={Advances in Neural Information Processing Systems},
  volume={35},
  pages={28940--28954},
  year={2022}
}

@article{zhao2025smoothing,
  title={Smoothing Slot Attention Iterations and Recurrences},
  author={Zhao, Rongzhen and Yang, Wenyan and Kannala, Juho and Pajarinen, Joni},
  journal={arXiv preprint arXiv:2508.05417},
  year={2025}
}

@inproceedings{karazija2clevrtex,
 author = {Karazija, Laurynas and Laina, Iro and Rupprecht, Christian},
 booktitle = {Proceedings of the Neural Information Processing Systems Track on Datasets and Benchmarks},
 editor = {J. Vanschoren and S. Yeung},
 pages = {},
 title = {ClevrTex: A Texture-Rich Benchmark for Unsupervised Multi-Object Segmentation},
 url = {https://datasets-benchmarks-proceedings.neurips.cc/paper_files/paper/2021/file/e2c420d928d4bf8ce0ff2ec19b371514-Paper-round2.pdf%7D},
 volume = {1},
 year = {2021},
}

@article{amir2021deep,
  title={Deep vit features as dense visual descriptors},
  author={Amir, Shir and Gandelsman, Yossi and Bagon, Shai and Dekel, Tali},
  journal={arXiv preprint arXiv:2112.05814},
  volume={2},
  number={3},
  pages={4},
  year={2021}
}

@article{rigas2025garlic,
  title={GARLIC: GAussian Representation LearnIng for spaCe partitioning},
  author={Rigas, Panagiotis and Drivas, Panagiotis and Tzamos, Charalambos and Chamodrakas, Ioannis and Ioannakis, George and Guibas, Leonidas J and Emiris, Ioannis Z},
  journal={arXiv preprint arXiv:2505.24608},
  year={2025}
}

@article{ueda2000smem,
  title={SMEM algorithm for mixture models},
  author={Ueda, Naonori and Nakano, Ryohei and Ghahramani, Zoubin and Hinton, Geoffrey E},
  journal={Neural computation},
  volume={12},
  number={9},
  pages={2109--2128},
  year={2000},
  publisher={MIT Press One Rogers Street, Cambridge, MA 02142-1209, USA journals-info~…}
}

@article{fritzke1994growing,
  title={A growing neural gas network learns topologies},
  author={Fritzke, Bernd},
  journal={Advances in neural information processing systems},
  volume={7},
  year={1994}
}

@article{kerbl20233d,
  title={3d gaussian splatting for real-time radiance field rendering.},
  author={Kerbl, Bernhard and Kopanas, Georgios and Leimk{\"u}hler, Thomas and Drettakis, George and others},
  journal={ACM Trans. Graph.},
  volume={42},
  number={4},
  pages={139--1},
  year={2023}
}

@InProceedings{Bock_2026_CVPR,
    author    = {Bock, Sebastian and Sch\"u{\ss}ler, Leonie and Singh, Krishnakant and Schaub-Meyer, Simone and Roth, Stefan},
    title     = {MUFASA: A Multi-Layer Framework for Slot Attention},
    booktitle = {Proceedings of the IEEE/CVF Conference on Computer Vision and Pattern Recognition (CVPR)},
    month     = {June},
    year      = {2026},
    pages     = {27750-27760}
}

@inproceedings{ouyang2026qasa,
  title={QASA: Quality-Guided K-Adaptive Slot Attention for Unsupervised Object-Centric Learning},
  author={Ouyang, Tianran and Dong, Xingping and Zhang, Jing and Ye, Mang and Zhang, Kaihao and Du, Bo},
  booktitle={European Conference on Computer Vision},
  pages={492--508},
  year={2026},
  organization={Springer}
}

@article{peters2021capturing,
  title={Capturing the objects of vision with neural networks},
  author={Peters, Benjamin and Kriegeskorte, Nikolaus},
  journal={Nature human behaviour},
  volume={5},
  number={9},
  pages={1127--1144},
  year={2021},
  publisher={Nature Publishing Group UK London}
}
}

\newpage
\appendix
\section{Supplementary Material}

The supplementary material presents the merging process and its observed behavior (Sec.~\ref{sec:supp_merging_process}), analyzes the design choices (Sec.~\ref{sec:supp_design_choices}), and reports extended experimental results (Sec.~\ref{sec:supp_extended_evaluation}). Experimental setup and implementation details are provided in Sec.~\ref{sec:supp_experimental_setup}, followed by downstream application protocols in Sec.~\ref{sec:supp_application_protocols}.

Unless otherwise stated, all experiments in the supplementary material integrate the proposed slot merging mechanism into DINOSAUR using the configuration described in Sec.~\ref{sec:train_conf}. Accordingly, \emph{Ours} denotes DINOSAUR with slot merging. Experiments using SPOT, SlotDiffusion, alternative feature extractors, or other configurations are explicitly identified.

\subsection{Understanding the Merging Process}
\label{sec:supp_merging_process}

\subsubsection{Iterative Merging Process}
Fig.~\ref{fig:merging_steps} demonstrates the merging steps for specific images starting from the representation that Slot Attention produces until the final step where the corresponding slots are passed to the decoder for further refinement. Each step merges the slots with the highest Soft-IoU score.

\begin{figure}[h]
    \centering
    \includegraphics[width=0.99\linewidth]{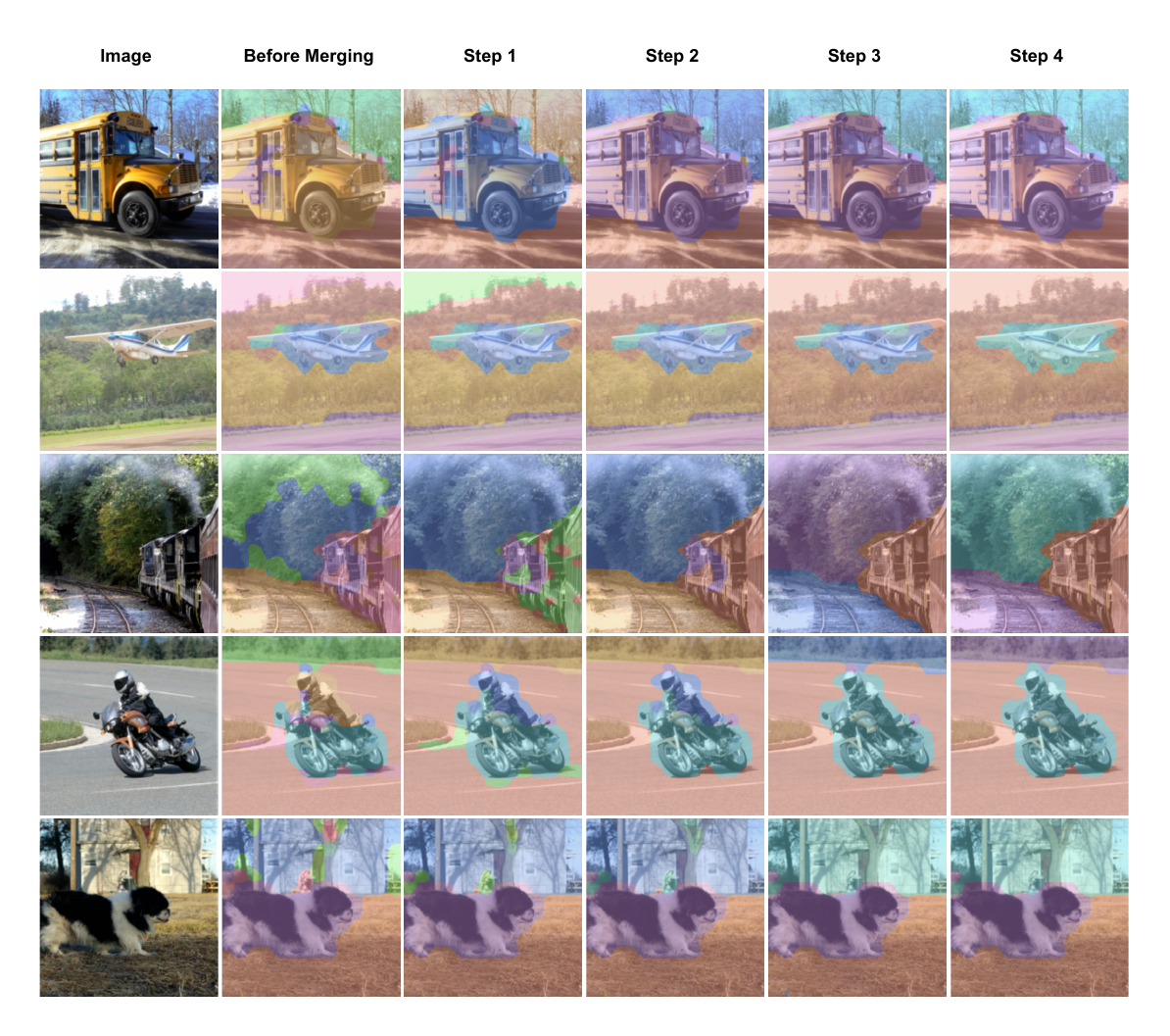}
    \caption{Visualizations of the iterative slot merging process. Starting from slot representations produced by Slot Attention, the algorithm repeatedly merges the pair of slots with the highest Soft-IoU overlap. The final merged slot set is then passed to the decoder for reconstruction.}
    \label{fig:merging_steps}
\end{figure}

\subsubsection{Merging Statistics}
To further analyze the behavior of the merging mechanism, Tab.~\ref{tab:aggregated_stats} reports aggregated merging statistics. The merge frequency per image remains consistent between training and inference, indicating stable operation. On COCO, which contains dense scenes with approximately 7.7 objects per image, the mean number of merges is lowest (0.69) despite the initialization $K=8$, suggesting that most slots are already specialized with limited overlap. In contrast, MOVi-E starts from $K=25$ slots, exceeding the typical object count, and exhibits substantially more merging (7.96 on average). MOVi-C shows a similar pattern: with $K=12$, the model performs about 2.2 merges per image. 

\begin{table}[ht]
\caption{Merging statistics during training and inference across all four datasets, reporting the mean and standard deviation of the number of merged slots per image.}
\label{tab:aggregated_stats}
\centering
\setlength{\tabcolsep}{4pt}
{%
\begin{tabular}{l @{\hskip 0.6em} cccc}
\toprule
& \multicolumn{4}{c}{Training / Inference} \\
\cmidrule(lr){2-5}
& VOC & COCO & MOVi-C & MOVi-E \\
\midrule
Mean & 1.67\;/\;1.73 & 0.69\;/\;0.82 & 2.2\;/\;2.2 & 7.96\;/\;7.91 \\
Std  & 1.0\;/\;0.98  & 0.76\;/\;0.81 & 1.32\;/\;1.32 & 2.13\;/\;2.08 \\
\bottomrule
\end{tabular}%
}
\end{table}

Inspired by \cite{fan2024adaptive}, we further compare the distribution of active slots after merging with the ground-truth object count on MOVi-E in Fig.~\ref{fig:slot_distribution}. Starting from a fixed $K=25$, our method produces a broader, image-dependent distribution of active slots that follows the variation in object density. The distribution is consistently shifted toward larger values, as active slots may additionally represent background regions, whereas the ground-truth count includes only foreground objects. 

\begin{figure}
    \centering
    \includegraphics[width=0.9\linewidth]{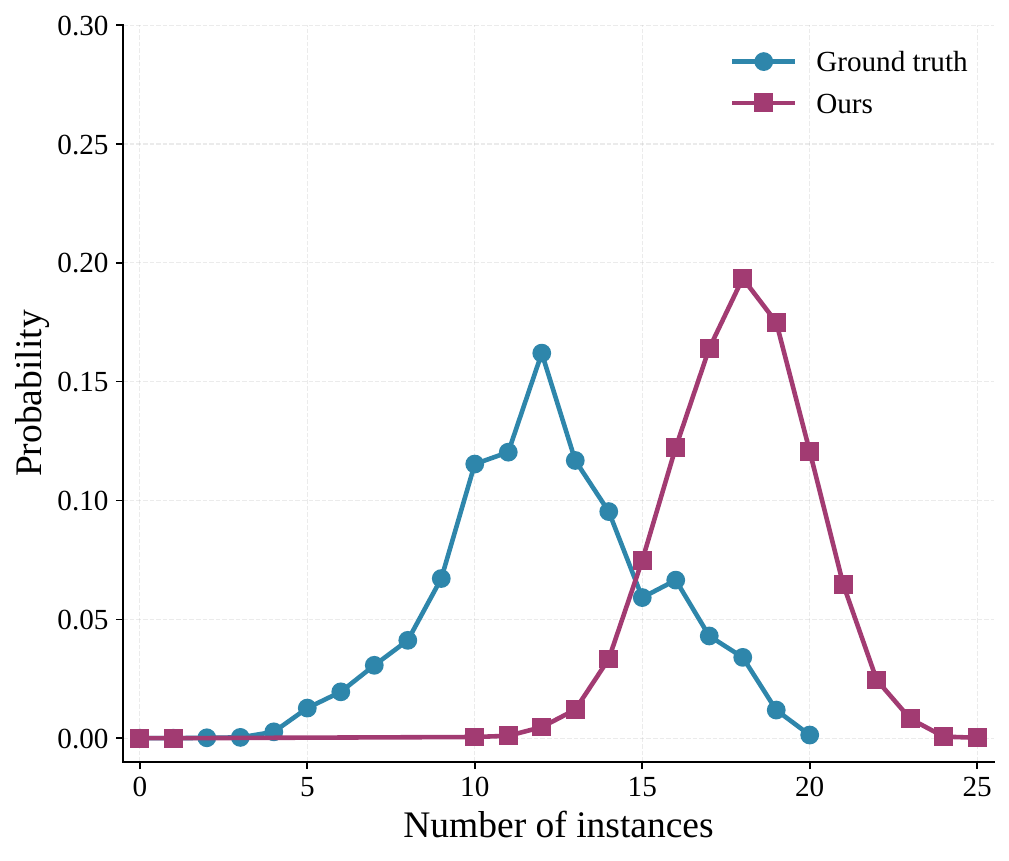}
    \caption{Distribution of ground-truth object count and active slots after merging on MOVi-E.}
    \label{fig:slot_distribution}
\end{figure}

\subsection{Analysis of Design Choices}
\label{sec:supp_design_choices}

\subsubsection{Threshold Selection and Sensitivity}
\label{sec:threshold_selection}

\noindent\textbf{Threshold Estimation.}
To regulate the iterative merging process, an adaptive stopping criterion is introduced. Merging is activated only once the training dynamics stabilize, ensuring that slot representations are sufficiently discriminative. Merging threshold $\tau$ is determined automatically by analyzing the distribution of pairwise Soft IoU scores over 10 batches, one epoch before merging starts. For each batch $b$, the IoU values are discretized into 100 bins to form a histogram $H_b$. To mitigate the influence of extreme counts and skewed distributions, a logarithmic transformation is applied.

Following~\cite{zack1977automatic}, a candidate threshold is extracted from histogram geometry. Specifically, a line is drawn between the histogram peak and farthest nonzero bin on the longer tail. The bin with the maximum perpendicular distance $d_{\perp}$ to this line separates high-overlap and low-overlap regions, and is selected as the batch-wise threshold candidate. To ensure robustness against batch-level variability, the candidate thresholds are aggregated using first-order statistics, such as mean or variance-adjusted mean. For VOC and MOVi-E, we select mean minus standard deviation among candidate thresholds, and mean value as threshold for COCO and MOVi-C. Selection is based on dataset statistics, leaving space for merging on datasets that can support it. Thresholding process is also depicted in Fig.~\ref{fig:triangle}.

\noindent\textbf{Threshold Sensitivity.}
Fig.~\ref{fig:variable_threshold} evaluates the effect of merging threshold on VOC dataset. Performance remains stable across a wide range of values, showing robustness. Both $mBO^i$ and $mBO^c$ exhibit only minor variations as threshold changes. Across all evaluated thresholds, the proposed method consistently outperforms DINOSAUR baseline.

\begin{figure}[t]
    \centering
    \begin{subfigure}{0.48\linewidth}
        \centering
        \includegraphics[width=\linewidth]{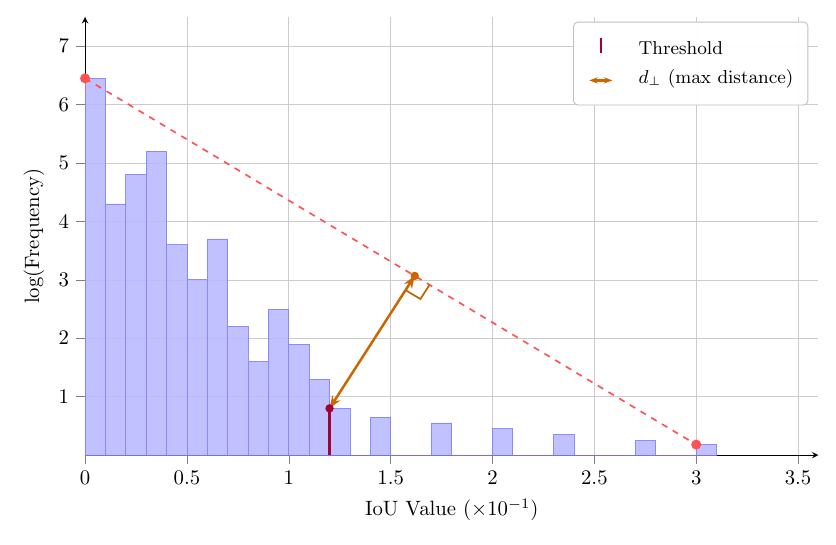}
        \caption{}
        \label{fig:triangle}
    \end{subfigure}
    \hfill
    \begin{subfigure}{0.48\linewidth}
        \centering
        \includegraphics[width=\linewidth]{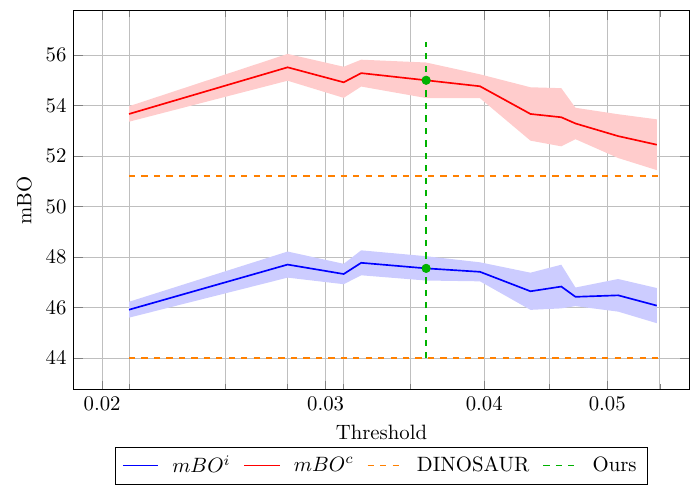}
        \caption{}
        \label{fig:variable_threshold}
    \end{subfigure}
    \caption{(a) Data-driven estimation of threshold from the histogram of pairwise Soft-IoU scores using the maximum perpendicular distance criterion. (b) Sensitivity analysis of the merging threshold on the VOC dataset measured using $mBO^i$ and $mBO^c$.}
    \label{fig:threshold_analysis}
\end{figure}

\subsubsection{Effect of Attention Map Update}
We examine whether updating the attention maps after merging contributes to performance. Specifically, attention map updates using Eq.~7 are compared against retaining the original attention map of a single slot after merging (Tab.~\ref{tab:attn_maps_fusion}). Removing attention addition leads to consistent degradation, suggesting that attention map fusion supports effective slot merging.

\begin{table}[b]
\caption{Effect of attention map aggregation during slot merging. Results are averaged over 3 seeds.}
\label{tab:attn_maps_fusion}
\centering
{
\begin{tabular}{l p{0.6em} ccc}
\toprule
& & mBO$^i$ & mBO$^c$ & mIoU \\
\midrule
w/o Aggregation  
& & 47.38{\tiny$\pm$0.52} & 54.83{\tiny$\pm$0.58} & 45.53{\tiny$\pm$0.49} \\

\rowcolor{oursGreen}
w/ Aggregation 
& & \bf{47.83{\tiny$\pm$0.17}} & \bf{55.34{\tiny$\pm$0.27}} & \bf{46.01{\tiny$\pm$0.18}} \\
\bottomrule
\end{tabular}
}
\end{table}

\subsubsection{Effect of Merging Activation Epoch}
Tab.~\ref{tab:merge_epoch} analyzes the effect of the epoch at which merging is introduced during training. On VOC, earlier activation yields the strongest performance, while delaying merging leads to consistent degradation. A similar trend is observed on MOVi-C, where early activation outperforms later introduction. The performance gap confirms that sufficient optimization time after enabling merging is critical. Earlier activation allows the model to adapt slot representations to the merging dynamics, whereas late activation restricts this adaptation window and limits the attainable gains.

\begin{table}[b]
\caption{Influence of the training epoch at which slot merging is activated, evaluated on VOC and MOVi-C.}
\label{tab:merge_epoch}
\centering
\setlength{\tabcolsep}{4pt}
{%
\begin{tabular}{l c ccc}
\toprule
& Epoch & mBO$^i$ & mBO$^c$ & mIoU \\
\midrule
\multirow{3}{*}{VOC}
& 90 & 46.15 & 52.91 & 44.65 \\
& 80 & 46.96 & 54.10 & 45.25 \\
& \cellcolor{oursGreen}70 & \cellcolor{oursGreen}\textbf{47.83} & \cellcolor{oursGreen}\textbf{55.33} & \cellcolor{oursGreen}\textbf{46.01} \\
\midrule
\multirow{3}{*}{MOVi-C}
& 50 & 47.05 & -- & 46.19 \\
& 40 & 46.83 & -- & 46.00 \\
& \cellcolor{oursGreen}30 & \cellcolor{oursGreen}\textbf{48.35} & \cellcolor{oursGreen}-- & \cellcolor{oursGreen}\textbf{47.43} \\
\bottomrule
\end{tabular}%
}
\end{table}

\subsection{Extended Experimental Evaluation}
\label{sec:supp_extended_evaluation}

\subsubsection{Comparison with Prior Object-Centric Methods}
In Tab.~\ref{tab:main_results_all}, we report extended benchmark results including both mBO and mIoU across four datasets for DINOSAUR~\cite{seitzer2022bridging} and SPOT~\cite{Kakogeorgiou_2024_CVPR} with our merging strategy. On MOVi-C, SPOT + Ours achieves the highest mBO among all compared methods, while on MOVi-E it achieves the highest mBO and mIoU. Our approach also achieves competitive performance compared with adaptive slot-number methods, demonstrating that refining an initially overcomplete set of slots through merging is an effective strategy for object discovery. Improvements on both DINOSAUR and SPOT support the use of slot merging as a complementary module across different architectures and training procedures.

\noindent\textbf{Inference-only Integration into GLASS.}
We also evaluate merging in the diffusion-based GLASS model~\cite{singh2025glass} on COCO, applying it after the final Slot Attention iteration without retraining. Under the original segmentation evaluation protocol, GLASS + Ours improves over the pretrained model and achieves the best results among all compared methods.

\begin{table*}[t]
\centering
\caption{Comparison with object-centric methods on VOC, COCO, MOVi-C, and MOVi-E datasets. Results for our approach are reported as mean $\pm$ standard deviation over 3 seeds. $\dag$ indicates results taken from~\cite{Kakogeorgiou_2024_CVPR}. Otherwise, results are from the original papers. The best results are shown in \textbf{bold} and the second-best results are \underline{underlined}.}
\label{tab:main_results_all}

\resizebox{\textwidth}{!}{
\begin{tabular}{l ccc ccc cc cc}
\toprule
& \multicolumn{3}{c}{VOC}
& \multicolumn{3}{c}{COCO}
& \multicolumn{2}{c}{MOVi-C}
& \multicolumn{2}{c}{MOVi-E} \\
\cmidrule(lr){2-4}
\cmidrule(lr){5-7}
\cmidrule(lr){8-9}
\cmidrule(lr){10-11}
Method
& mBO$^i$ & mBO$^c$ & mIoU
& mBO$^i$ & mBO$^c$ & mIoU
& mBO$^i$ & mIoU
& mBO$^i$ & mIoU \\
\midrule

SA \cite{locatello2020object}$^{\dag}$ {\tiny NeurIPS'20} &
24.6 & 24.9 & -- &
17.2 & 19.2 & -- &
26.2{\tiny$\pm$1.0} & -- &
24.0{\tiny$\pm$1.2} & -- \\

SLATE \cite{singh2021illiterate}$^{\dag}$ {\tiny ICLR'22} &
35.9 & 41.5 & -- &
29.1 & 33.6 & -- &
39.4{\tiny$\pm$0.8} & 37.8{\tiny$\pm$0.7} &
30.2{\tiny$\pm$1.7} & 28.6{\tiny$\pm$1.7} \\

SLASH \cite{kim2023shepherding}$^{\dag}$ {\tiny CVPR'23} &
-- & -- & -- &
-- & -- & -- &
-- & 27.7{\tiny$\pm$5.9} &
-- & -- \\

CAE \cite{lowecomplex}$^{\dag}$ {\tiny TMLR} &
32.9{\tiny$\pm$0.9} & 37.4{\tiny$\pm$1.0} & -- &
-- & -- & -- &
-- & -- &
-- & -- \\

Top-Down Slot Attention \cite{kim2024bootstrapping} {\tiny NeurIPS'24} &
43.9{\tiny$\pm$2.6} & 51.0{\tiny$\pm$2.5} & 42.0{\tiny$\pm$2.8} &
33.0{\tiny$\pm$0.3} & 40.3{\tiny$\pm$0.2} & 31.2{\tiny$\pm$0.3} &
46.8{\tiny$\pm$2.4} & 45.9{\tiny$\pm$2.5} &
39.3{\tiny$\pm$1.8} & 38.3{\tiny$\pm$1.9} \\

Rotating Features \cite{lowe2023rotating}$^{\dag}$ {\tiny NeurIPS'23} &
40.7{\tiny$\pm$0.1} & 46.0{\tiny$\pm$0.1} & -- &
-- & -- & -- &
-- & -- &
-- & -- \\

DIAS$_{\text{image}}$ \cite{zhao2025slot} {\tiny ACM MM'25} &
44.8{\tiny$\pm$9.8} & -- & 42.8{\tiny$\pm$1.7} &
32.8{\tiny$\pm$0.1} & -- & 30.1{\tiny$\pm$0.3} &
-- & -- &
-- & -- \\

\midrule
\multicolumn{11}{l}{\textbf{DINOSAUR-Based}} \\

DINOSAUR-MLP \cite{seitzer2022bridging}$^{\dag}$ {\tiny ICLR'23} &
39.5{\tiny$\pm$0.1} & 40.9{\tiny$\pm$0.1} & -- &
27.7{\tiny$\pm$0.2} & 30.9{\tiny$\pm$0.2} & -- &
39.1{\tiny$\pm$0.2} & -- &
35.5{\tiny$\pm$0.2} & -- \\

DINOSAUR \cite{seitzer2022bridging}$^{\dag}$ {\tiny ICLR'23} &
44.0{\tiny$\pm$1.9} & 51.2{\tiny$\pm$1.9} & -- &
32.3{\tiny$\pm$0.4} & 38.8{\tiny$\pm$0.4} & -- &
42.4 & -- &
-- & -- \\

DINOSAUR+FS+RC \cite{tian2025pay} {\tiny CVPR'25} &
45.2 & 52.9 & -- &
31.2 & 40.1 & -- &
46.0 & 44.8 &
-- & -- \\

DINOSAUR-M \cite{Bock_2026_CVPR} {\tiny CVPR'26} &
49.2{\tiny$\pm$0.4} & 57.6{\tiny$\pm$0.6} & 47.2{\tiny$\pm$0.5} &
32.7{\tiny$\pm$0.4} & 43.0{\tiny$\pm$0.7} & 30.5{\tiny$\pm$0.3} &
\underline{49.2{\tiny$\pm$0.5}} & \textbf{48.3{\tiny$\pm$0.5}} &
-- & -- \\

\midrule
\multicolumn{11}{l}{\textbf{Autoregressive Teacher-Student}} \\

SPOT \cite{Kakogeorgiou_2024_CVPR} {\tiny CVPR'24} &
48.3{\tiny$\pm$0.4} & 55.6{\tiny$\pm$0.4} & 46.8{\tiny$\pm$0.4} &
35.0{\tiny$\pm$0.1} & 44.7{\tiny$\pm$0.3} & 33.0{\tiny$\pm$0.1} &
47.3{\tiny$\pm$1.2} & 46.7{\tiny$\pm$1.3} &
\underline{40.1{\tiny$\pm$1.2}} & \underline{39.3{\tiny$\pm$1.2}} \\

SPOT+FS+RC \cite{tian2025pay} {\tiny CVPR'25} &
49.3 & 56.5 & -- &
35.7 & 45.3 & -- &
49.0 & 47.8 &
-- & -- \\

SPOT-M \cite{Bock_2026_CVPR} {\tiny CVPR'26} &
\underline{51.3{\tiny$\pm$0.1}} &
\underline{59.8{\tiny$\pm$0.3}} &
\underline{49.4{\tiny$\pm$0.2}} &
34.8{\tiny$\pm$0.2} &
45.5{\tiny$\pm$0.4} &
32.5{\tiny$\pm$0.2} &
\underline{49.2{\tiny$\pm$0.3}} &
\underline{48.2{\tiny$\pm$0.3}} &
-- & -- \\

\midrule
\multicolumn{11}{l}{\textbf{Diffusion}} \\

SlotDiffusion \cite{wu2023slotdiffusion}$^{\dag}$ {\tiny NeurIPS'23} &
50.4 & 55.3 & -- &
31.0 & 35.0 & -- &
-- & -- &
30.2 & 30.2 \\

LSD \cite{jiang2023object}$^{\dag}$ {\tiny NeurIPS'23} &
-- & -- & -- &
-- & -- & -- &
45.6{\tiny$\pm$0.8} & 44.2{\tiny$\pm$0.9} &
39.0{\tiny$\pm$0.5} & 37.6{\tiny$\pm$0.5} \\

Stable-LSD \cite{jiang2023object}$^{\dag}$ {\tiny NeurIPS'23} &
-- & -- & -- &
30.4 & -- & -- &
-- & -- &
-- & -- \\

GLASS \cite{singh2025glass} {\tiny CVPR'25} &
\textbf{58.9} & \textbf{62.2} & \textbf{58.1} &
\underline{40.8} & \underline{48.7} & \underline{39.0} &
-- & -- &
-- & -- \\

\midrule
\multicolumn{11}{l}{\textbf{Adaptive Slot Attention}} \\

AdaSlot \cite{fan2024adaptive} {\tiny CVPR'24} &
-- & -- & -- &
27.4 & -- & -- &
35.6 & -- &
29.8 & -- \\

MetaSlot-MLP \cite{liu2025metaslot} {\tiny NeurIPS'25} &
43.9{\tiny$\pm$0.3} & -- & 42.1{\tiny$\pm$0.2} &
29.5{\tiny$\pm$0.2} & -- & 27.9{\tiny$\pm$0.2} &
35.0 & -- &
-- & -- \\

MetaSlot-Tfd \cite{liu2025metaslot} {\tiny NeurIPS'25} &
39.0{\tiny$\pm$0.3} & -- & 37.8{\tiny$\pm$0.5} &
28.2{\tiny$\pm$0.7} & -- & 26.7{\tiny$\pm$0.6} &
-- & -- &
-- & -- \\

MetaSlot-Dfz \cite{liu2025metaslot} {\tiny NeurIPS'25} &
36.5{\tiny$\pm$0.1} & -- & 35.3{\tiny$\pm$0.1} &
27.2{\tiny$\pm$0.2} & -- & 25.8{\tiny$\pm$0.1} &
-- & -- &
-- & -- \\

QASA \cite{ouyang2026qasa} {\tiny ECCV'26} &
49.7 & 57.9 & 47.9 &
36.7 & 45.5 & 35.0 &
46.9 & 46.1 &
39.1 & 37.8 \\

QASA$_{\text{MLP}}$ \cite{ouyang2026qasa} {\tiny ECCV'26} &
47.3 & 53.8 & 45.5 &
33.6 & 36.5 & 32.5 &
42.2 & 41.1 &
34.6 & 33.0 \\

\midrule
\rowcolor{oursGreen}
\textbf{DINOSAUR + Ours} &
47.83{\tiny$\pm$0.17} &
55.34{\tiny$\pm$0.27} &
46.01{\tiny$\pm$0.18} &
32.88{\tiny$\pm$0.40} &
42.78{\tiny$\pm$0.22} &
30.75{\tiny$\pm$0.55} &
48.35{\tiny$\pm$1.32} &
47.43{\tiny$\pm$1.26} &
39.56{\tiny$\pm$0.55} &
37.75{\tiny$\pm$0.55} \\

\rowcolor{oursGreen}
\textbf{SPOT + Ours} &
49.44{\tiny$\pm$0.63} &
57.35{\tiny$\pm$0.66} &
47.62{\tiny$\pm$0.62} &
34.77{\tiny$\pm$0.41} &
45.18{\tiny$\pm$0.35} &
32.54{\tiny$\pm$0.46} &
\textbf{49.50{\tiny$\pm$0.52}} &
48.17{\tiny$\pm$0.83} &
\textbf{41.39{\tiny$\pm$0.42}} &
\textbf{39.94{\tiny$\pm$0.45}} \\

\rowcolor{oursGreen}
\textbf{GLASS + Ours} &
-- &
-- &
-- &
\textbf{41.8} &
\textbf{50.6} &
\textbf{39.7} &
-- &
-- &
-- &
-- \\

\bottomrule
\end{tabular}
}
\end{table*}

\subsubsection{Comparison with other Pre-trained Feature Extractors}
In our experiments, we use DINO features as training signal. We also evaluate DINOv2~\cite{oquab2023dinov2} and MAE~\cite{he2022masked} as alternative feature extractors on VOC. As shown in Tab.~\ref{tab:encoder_comparison}, our method improves over DINOSAUR baseline across all encoders, with the largest gains observed when using DINO features.

\begin{table}[t]
\caption{Different encoders comparison. DINO~\cite{caron2021emerging} and MAE \cite{he2022masked} use ViT-B/16 architecture~\cite{dosovitskiy2020image} while DINOv2~\cite{oquab2023dinov2} employs ViT-S/14.}
\label{tab:encoder_comparison}
\centering
\resizebox{0.5\textwidth}{!}{
\begin{tabular}{l p{0.6em} l p{0.8em} c c c}
\toprule
Encoder && Method && mBO$^i$ & mBO$^c$ & mIoU \\
\midrule

\multirow{2}{*}{DINO}
    && DINOSAUR && 44.0 & 51.2 & -- \\
    & \cellcolor{oursGreen} & \cellcolor{oursGreen}Ours & \cellcolor{oursGreen} & \cellcolor{oursGreen}47.8 & \cellcolor{oursGreen}55.3 & \cellcolor{oursGreen}46.0 \\

\midrule

\multirow{2}{*}{DINOv2}
    && DINOSAUR && 43.8 & 48.1 & 42.4 \\
    & \cellcolor{oursGreen} & \cellcolor{oursGreen}Ours & \cellcolor{oursGreen} & \cellcolor{oursGreen}45.3 & \cellcolor{oursGreen}51.3 & \cellcolor{oursGreen}43.7 \\

\midrule

\multirow{2}{*}{MAE}
    && DINOSAUR && 37.8 & 42.7 & 36.5 \\
    & \cellcolor{oursGreen} & \cellcolor{oursGreen}Ours & \cellcolor{oursGreen} & \cellcolor{oursGreen}38.3 & \cellcolor{oursGreen}43.4 & \cellcolor{oursGreen}36.7 \\

\bottomrule
\end{tabular}
}
\end{table}

\subsubsection{Behavior under Different Object Sizes and Occlusions}
The analysis with respect to object size (Fig.~\ref{fig:mboi_instancesize}) shows that the remaining performance gap is primarily concentrated on very small instances, where object discovery is already challenging for the baseline, while consistent improvements are observed for medium and large objects.

\begin{figure}[ht]
    \centering
    
    \begin{subfigure}{0.5\textwidth}
        \centering
        \includegraphics[width=\linewidth]{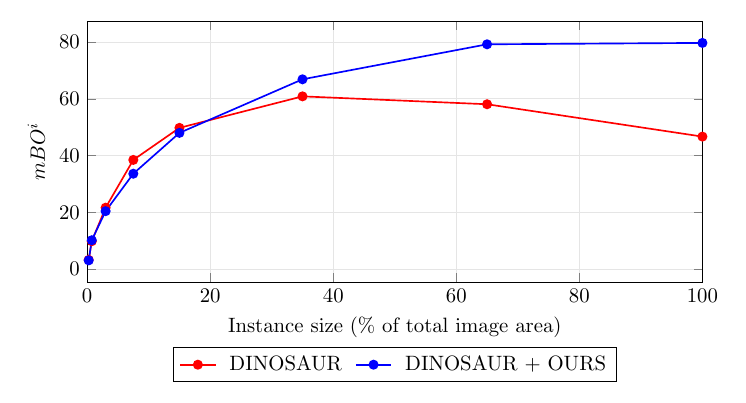}
        \label{fig:first}
    \end{subfigure}
    \hfill
    \begin{subfigure}{0.5\textwidth}
        \centering
        \includegraphics[width=\linewidth]{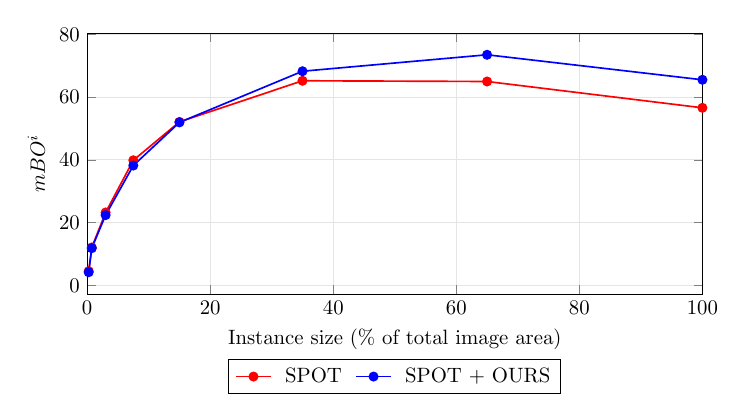}
        \label{fig:second}
    \end{subfigure}
    \caption{Analysis of DINOSAUR (top) and SPOT (bottom) with and w/o merging performance across different instance sizes. The performance is measured in $mBO^i$ on VOC, across varied instance sizes. The instance sizes are expressed as a percentage of the total image area, categorized into distinct bins: 0--0.5\%, 0.5--1\%, 1--5\%, 5--10\%, 10--20\%, 20--50\%, 50--80\%, 80--100\%.}
    \label{fig:mboi_instancesize}
\end{figure}

Occluded scenes present a challenging case for the proposed approach, particularly when neighboring or partially occluded objects are not sufficiently separated in the underlying DINO feature space. Since the proposed merging operates on the Slot Attention maps derived from these frozen representations, ambiguities in the feature embeddings may result in overlapping slot assignments and consequently incorrect merging decisions. This behavior is inherited from the backbone representation rather than introduced by the proposed merging strategy. In particular, DINO features are known to encode strong semantic correspondences but do not explicitly enforce instance-level separation~\cite{caron2021emerging,amir2021deep}. Nevertheless, the occlusion analysis in Fig.~\ref{fig:occlusions} shows that improvements over DINOSAUR are maintained across a broad range of occlusion levels, suggesting that the proposed approach remains effective despite the potential ambiguity associated with highly similar feature representations.

\begin{figure}
    \centering
    \includegraphics[width=0.59\linewidth]{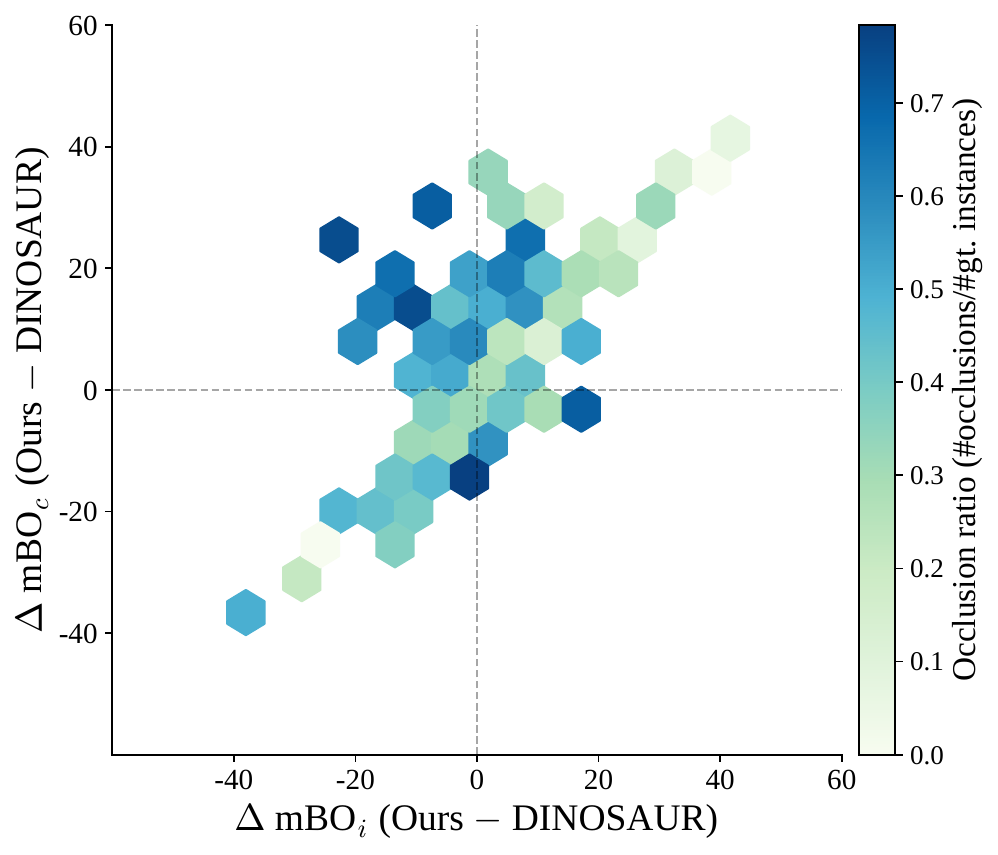}
    \caption{Occlusion analysis on VOC dataset. Axes show performance gains in terms of mBO for instance ($i$) and class ($c$) metrics. Color denotes the occlusion ratio per hexagonal bin.}
    \label{fig:occlusions}
\end{figure}

\subsubsection{Training Time}
We measure training time for VOC and MOVi-C datasets. Training time is defined as the time spent within the training loop, including forward pass, backward pass, gradient updates, and data loading. Tab.~\ref{tab:training_time} reports total training time on the two datasets. As shown, on VOC, the difference between DINOSAUR and our method is negligible, indicating that the proposed modification introduces only a minimal practical computational overhead on datasets with a relatively small number of slots $K$. On MOVi-C, the runtime increases slightly, but the overall overhead remains small.

\begin{table}[t]
\centering
\caption{Training time comparison between DINOSAUR and our method.}
\label{tab:training_time}
\resizebox{0.5\textwidth}{!}{
\begin{tabular}{l p{0.8em} c p{0.8em} c}
\toprule
Method && VOC && MOViC \\
\midrule
DINOSAUR && $\sim$ 53 minutes  && $\sim$ 4h 44 minutes \\
\rowcolor{oursGreen}
Ours && $\sim$ 53.5 minutes && $\sim$ 4h 50 minutes \\
\bottomrule
\end{tabular}
}
\end{table}

\subsection{Experimental Setup and Implementation}
\label{sec:supp_experimental_setup}

\subsubsection{Datasets and Preprocessing}
\noindent\textbf{PASCAL VOC.}
We train on PASCAL VOC 2012 trainaug set (10,582 images), which combines 1,464 images from the segmentation train set and 9,118 images from the SBD dataset~\cite{hariharan2011semantic}. This split is used following prior work~\cite{Kakogeorgiou_2024_CVPR,seitzer2022bridging}. During training, the image minor axis is resized to 224 pixels, followed by random cropping to $224 \times 224$ and random horizontal flipping with probability 0.5. Evaluation is performed on the official segmentation validation set (1,449 images), ignoring unlabeled pixels. We resize the minor axis to 320 pixels, apply center cropping, and evaluate the masks at $320 \times 320$ resolution.

\noindent\textbf{COCO.}
We train on the COCO 2017 dataset (118,287 images) and evaluate our method on the validation set (5,000 images). During training, the image minor axis is resized to $224 \times 224$, followed by a $224 \times 224$ center crop and random horizontal flipping with probability 0.5. For evaluation, the minor axis is resized to 320 pixels, center cropped, and masks are evaluated at a resolution of $320 \times 320$ as described in~\cite{Kakogeorgiou_2024_CVPR,seitzer2022bridging}.

\noindent\textbf{MOVi-C/E.}
We convert the MOVi-C/E video datasets into image datasets by sampling nine random frames from each training clip, resulting in 87,633 images for MOVi-C and 87,741 for MOVi-E. During training, images are resized to $224 \times 224$. For evaluation, we use all frames from the 250 validation clips (6,000 images), following prior work~\cite{Kakogeorgiou_2024_CVPR,seitzer2022bridging}, and evaluate the masks at the full $128 \times 128$ resolution.

\subsubsection{Decoder Integration and Masking}
Alg.~1 returns the updated slot tensor $S$ and the binary mask $M$, where $M_k=1$ indicates an active slot. For each image, the active slot set is $\mathcal{S}=\{S_k\mid M_k=1\}$. All integrations retain the original tensor shape, while merged-away slots are excluded from decoder cross-attention using $M$. Specifically, inactive slots are masked before the softmax and therefore receive zero attention weight. This is necessary because setting a slot vector to zero would still retain its key in the attention normalization. For DINOSAUR~\cite{seitzer2022bridging} and SPOT~\cite{Kakogeorgiou_2024_CVPR}, the mask is applied to patch-to-slot cross-attention in every autoregressive decoder block. For SlotDiffusion~\cite{wu2023slotdiffusion} and GLASS~\cite{singh2025glass}, the same mask is applied to every slot-conditioned cross-attention block across all resolutions and denoising steps. All remaining decoder operations stay unchanged.

\subsubsection{Training Configuration}
\label{sec:train_conf}

We train all models using the Adam optimizer with $\beta_1=0.9$, $\beta_2=0.99$, no weight decay, and a batch size of 64. The learning rate is linearly warmed up from zero to its peak value over the first 10,000 training iterations, followed by cosine annealing to a predefined minimum value. For experiments using DINO~\cite{caron2021emerging}, the peak and minimum learning rates are set to $4\times10^{-4}$ and $4\times10^{-7}$, respectively, on VOC~\cite{everingham2012pascal} and COCO~\cite{lin2014microsoft}, and to $2\times10^{-4}$ and $4\times10^{-5}$ on MOVi-C and MOVi-E~\cite{greff2022kubric}. For experiments using DINOv2~\cite{oquab2023dinov2} or MAE~\cite{he2022masked} on VOC, the peak and minimum learning rates are set to $2\times10^{-4}$ and $4\times10^{-5}$, respectively. Slots are initialized from a shared Gaussian distribution, and gradients are clipped to an infinity norm of $0.3$.

Following the architectural configurations of DINOSAUR~\cite{seitzer2022bridging} and SPOT~\cite{Kakogeorgiou_2024_CVPR}, all models comprise a ViT-B/16 encoder~\cite{dosovitskiy2020image} and an autoregressive transformer decoder with four blocks and six attention heads per block. Slot Attention uses three refinement iterations and a slot dimension of 256, with merging applied after the third iteration. The model is initialized with 7, 8, 12, and 25 slots on VOC, COCO, MOVi-C, and MOVi-E, respectively.

DINOSAUR is trained for 100 epochs on VOC and COCO and 65 epochs on MOVi-C and MOVi-E. Merging is enabled from epoch 70 on VOC and COCO and from epoch 30 on MOVi-C and MOVi-E. We use thresholds of $\tau=0.036$, $0.04$, $0.035$, and $0.019$ for VOC, COCO, MOVi-C, and MOVi-E, respectively. These values are estimated from the overlap statistics as described in Sec.~\ref{sec:threshold_selection} and remain fixed after merging is enabled. All experiments are conducted on a single NVIDIA RTX A6000 GPU with 48\,GB of memory.

\subsubsection{SPOT Configuration}
\label{sec:spot_conf}

SPOT follows its original two-stage teacher-student training procedure~\cite{Kakogeorgiou_2024_CVPR}. The teacher is trained for 560 epochs on VOC, 50 epochs on COCO, and 65 epochs on MOVi-C and MOVi-E. Student training then continues for an additional 560, 50, and 30 epochs on VOC, COCO, and both MOVi datasets, respectively. For student training on MOVi-C and MOVi-E, the main and minimum learning rates are set to $2\times10^{-4}$ and $1.5\times10^{-4}$, respectively. During teacher training, merging is enabled from epochs 500, 40, 30, and 30 on VOC, COCO, MOVi-C, and MOVi-E, respectively, using the same dataset-specific thresholds as DINOSAUR.

During distillation, the teacher remains frozen and continues to apply slot merging, providing the resulting attention maps for Hungarian matching with the student predictions. The student's slot decomposition remains unchanged during optimization to preserve a stable distillation target. Merging is therefore applied to the student only at inference, using thresholds of $\tau=0.11$, $0.099$, $0.08$, and $0.065$ on VOC, COCO, MOVi-C, and MOVi-E, respectively.

\subsection{Downstream Application Protocols}
\label{sec:supp_application_protocols}

\subsubsection{Set Prediction}
Following standard object-property prediction protocols~\cite{seitzer2022bridging}, a shared lightweight predictor maps each active slot to class logits over the $20$ PASCAL VOC categories and background, together with the normalized center coordinates of the corresponding object. Since slots and ground-truth objects are unordered, predictions are matched to objects using the Hungarian algorithm with a cost combining class prediction and squared center distance. The predictor is trained using cross-entropy for classification and mean squared error for localization. We report averaged top-1 and top-5 class accuracy over foreground categories, together with the $R^2$ score for the predicted $x$ and $y$ coordinates.

\subsubsection{Object Recognition}
A shared lightweight head maps each active slot to class logits and a normalized bounding box. Unlike set prediction, slot to object correspondence is determined from the Slot Attention maps rather than from the head predictions. Following~\cite{zhao2025smoothing}, we upsample the attention maps to the ground-truth resolution, assign each pixel to its highest-attention active slot, and match the resulting slot masks to ground-truth instances using Hungarian matching on mask IoU. Matched pairs with IoU below $0.1$ are discarded. The head is trained using cross-entropy for classification and an $\ell_1$ loss for bounding-box regression. We report averaged top-1, top-3, and top-5 class accuracy, mean bounding-box IoU, and the ratio of matched ground-truth objects.

\subsubsection{Compositional Generation}
We evaluate compositional generation using a pretrained SlotDiffusion model. Following its standard protocol~\cite{wu2023slotdiffusion}, slots are cyclically exchanged across images in a batch, so each generated scene combines slots from different source images. The corresponding binary masks are also exchanged, so that slots removed by merging remain inactive during generation.

\subsubsection{Slot-Level Scene Editing}
We further evaluate direct manipulation of the active slot set $\mathcal{S}$. Target-object slots $\mathcal{S}_o$ are selected according to their overlap with the target mask~\cite{wu2023slotdiffusion}. The background set $\mathcal{S}_{\mathrm{bg}}$ contains the largest-mass active slot together with small-mass support slots, which consistently affect background appearance in our visualizations. The same selection rule is used for all examples. Removal decodes $\mathcal{S}\setminus\mathcal{S}_o$, while segmentation decodes $\mathcal{S}_o\cup\mathcal{S}_{\mathrm{bg}}$. Insertion adds source object slots to the target scene and decodes $\mathcal{S}^{A}\cup\mathcal{S}^{B}_o$. Background generation decodes only $\mathcal{S}_{\mathrm{bg}}$, while background swapping decodes $\mathcal{S}_{\mathrm{fg}}^{A}\cup\mathcal{S}_{\mathrm{bg}}^{B}$, where $A$ is the target image, $B$ is the source image.

\end{document}